\documentclass[lettersize,journal]{IEEEtran}

\ifCLASSINFOpdf
  \usepackage[pdftex]{graphicx}
\else
\fi
\usepackage[utf8]{inputenc}
\usepackage{amsmath,amsfonts}
\usepackage{dblfloatfix}
\usepackage{url}
\usepackage{xcolor}
\usepackage{enumerate}
\usepackage{xspace}
\usepackage{wrapfig}
\usepackage{comment}
\usepackage{multirow}

\usepackage{subcaption}

\begin{document}

\title{TrainSim: A Railway Simulation Framework for LiDAR and Camera Dataset Generation}


\author{
    \IEEEauthorblockN{
    Gianluca D'Amico\IEEEauthorrefmark{1},
    Mauro Marinoni\IEEEauthorrefmark{1},
    Federico Nesti\IEEEauthorrefmark{1},
    Giulio Rossolini\IEEEauthorrefmark{1},
    Giorgio Buttazzo\IEEEauthorrefmark{1}},
    Salvatore Sabina\IEEEauthorrefmark{1}\IEEEauthorrefmark{2},
    Gianluigi Lauro\IEEEauthorrefmark{2},
    \IEEEauthorblockA{\IEEEauthorrefmark{1}\textit{Department of Excellence in Robotics \& AI}, Scuola Superiore Sant'Anna, Pisa, Italy}
    \\\IEEEauthorblockA{\IEEEauthorrefmark{2}\textit{Hitachi Rail STS}, Genova, Italy}
    \thanks{This work has been submitted to the IEEE for possible publication. Copyright may be transferred without notice, after which this version may no longer be accessible.}
}



\maketitle

\begin{abstract}
The railway industry is searching for new ways to automate a number of complex train functions, such as object detection, track discrimination, and accurate train positioning, which require the artificial perception of the railway environment through different types of sensors, including cameras, LiDARs, wheel encoders, and inertial measurement units.
A promising approach for processing such sensory data is the use of deep learning models, which proved to achieve excellent performance in other application domains, as robotics and self-driving cars.
However, testing new algorithms and solutions requires the availability of a large amount of labeled data, acquired in different scenarios and operating conditions, which are difficult to obtain in a real railway setting due to strict regulations and practical constraints in accessing the trackside infrastructure and equipping
a train with the required sensors.
To address such difficulties, this paper presents a visual simulation framework able to generate realistic railway scenarios in a virtual environment and automatically produce inertial data and labeled datasets from emulated LiDARs and cameras useful for training deep neural networks or testing innovative algorithms.
A set of experimental results are reported to show the effectiveness of the proposed approach.
\end{abstract}



\section{Introduction} \label{s:intro}

\IEEEPARstart{I}{ncreasing} the efficiency and the safety of the railway network requires the automation of several complex functions, as signal  recognition, object detection, track discrimination, as well as an accurate train positioning. 
Most of these functions imply the execution of sophisticated perceptual tasks that must process and integrate data in real time from different heterogeneous sensors, as cameras, LiDARs, inertial measurement units (IMUs), wheel encoders, the global navigation satellite system (GNSS) receivers, and the transponders placed along the track line (balises).

Current train position functions are mainly based on balises, which include two components: a transponder on the track and a balise reader installed on the on-board train subsystem. When the reader detects a balise transponder on the track, the train position function is able to estimate the absolute position of the train.
Between consecutive balises, the relative position of the train is estimated via odometry by integrating the rotation speed measured by wheel encoders. Due to the integration operation and slip and slide phenomena that can occur between the wheels and the rails, the estimated train position is normally affected by a large drift, which causes the position error to increase with the travelled distance. Moreover, balises have a high deployment and maintenance cost and are subject to tampering.

Railway stakeholders are aiming at increasing the capacity of the line and reducing setup phases to provide cost-effective solutions capable of improving the exploitation of existing infrastructures, as demonstrated by the activities of the ERTMS User Group~\cite{EUGTLS}. However, reaching these goals requires improvements in the track discrimination function and a more precise train localization function in terms of odometry.

In the last years, to overcome the limitations of odometry algorithms based on balises and wheel encoders, different navigation solutions have been proposed to integrate GNSS data with inertial measurements through a Kalman filter~\cite{mirabadi1996application}.
Unfortunately, however, the GNSS signal is not always available (e.g., in tunnels and canyons) and is affected by multi-path phenomena that reduce the accuracy of its positioning. On the other hand, inertial data need to be integrated to estimate the position, and the integration process leads to increasing drifts over the time. Moreover, the high vibrations present in a railway environment have to be carefully addressed to limit and bound the position errors.

To overcome such problems, camera and LiDAR odometry algorithms have been integrated in navigation systems to reduce the localization error in different scenarios~\cite{tschopp2019experimental}.
Cameras and LiDAR sensors are also essential for other tasks related to signal recognition, object detection, and track discrimination, which consists of identifying the track where the train is running with respect to the other tracks present on the line. 
In particular, LiDARs are preferred to cameras, since the produced data are less affected by illumination conditions (due to the active nature of the LiDAR sensor) and because they directly provide the distances between the objects and the sensor in the form of point cloud.

Given the great performance of machine learning algorithms in several perceptual tasks, deep neural networks are the first candidates to be used for all the tasks mentioned above. 
More recently, several neural models have also been proposed to process the point cloud generated by a LiDAR~\cite{garcia2017review},~\cite{zhang2019review}, as well as to integrate camera and LiDAR data~\cite{caltagirone2019lidar},~\cite{asvadi2018multimodal}.

Unlike the automotive domain, where it is possible to find several studies and open datasets (e.g., Kitti~\cite{KITTI}) among the scientific community, the major obstacle in using deep learning algorithms in a railway environment is the lack of available datasets for training and testing new models. 
Also, planning an acquisition campaign is much more complex than in the automotive scenario, due to strict regulations and practical constraints in accessing the trackside infrastructure and equipping a train with the required sensors.

To address this issue, this paper proposes TrainSim, a simulation framework that generates synthetic datasets for training and testing deep neural networks, processing images, point clouds, and inertial data in a variety of railway environments and operating conditions.

In particular, the paper provides the following contributions:
\begin{enumerate}
    \item It presents a highly configurable and extendable environment generator to create a wide range of realistic railway scenarios controlled by a set of user-defined parameters.
    \item It proposes a method for generating arbitrarily large labeled datasets from such virtual environments using a set of simulated sensors (LiDARs, cameras, and IMUs) that can produce data similar to their physical counterparts.
    \item It provides a method for exporting the obtained datasets in a standard format for training deep neural networks or streaming them to ROS~\cite{ros} for real-time visualization.
\end{enumerate}

\begin{figure}[htbp!]
\centering
\makebox[\columnwidth]{\includegraphics[scale=0.3]{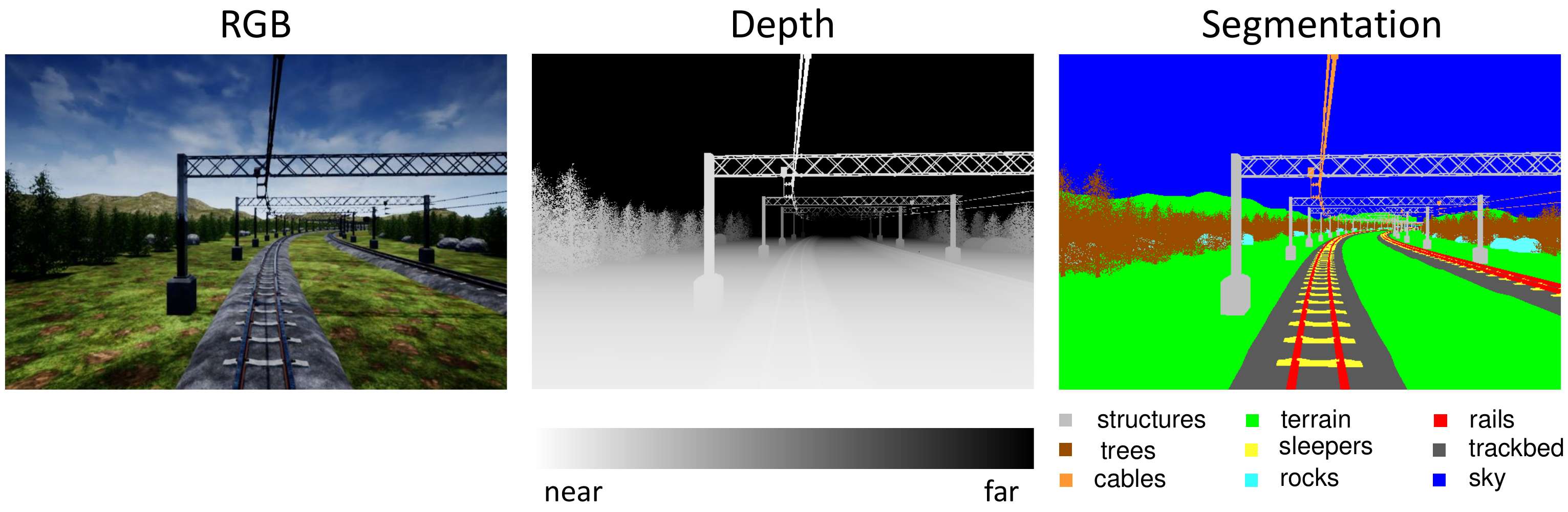}}
\vspace{-5mm}
\caption{Examples of images from the same scene produced by the simulator. Left: RGB camera; center: depth camera (each pixel value encodes the distance between the object represented by the pixel and the camera); right: segmented image (each pixel color encodes a different object class).}
\label{fig:images-example}
\end{figure}

At present, the proposed simulation framework, TrainSim, can generate the following datasets:
\begin{itemize}
    \item RGB images: taken from one or multiple cameras placed on the front vehicle of the train, in the positions specified by the user. 
    \item Depth images: taken from one or multiple cameras placed in user-defined positions, where each pixel value encodes the distance between the camera and the object point corresponding to the pixel. They can be used as ground truth data for depth estimation algorithms.
    \item Segmented images: taken from one or more cameras placed in the same position of the RGB or Depth cameras, where each pixel value encodes the class of the object corresponding to the pixel. They are used as labels for semantic segmentation and other tasks.
    \item Point Clouds: taken from a LiDAR sensor placed on the locomotive in a position specified by the user. A point cloud includes a set of 3D points acquired according to the scanning pattern of the user-defined LiDAR configuration.
    \item Segmented Point Clouds: taken from the same LiDAR. Each point is associated with a label that identifies the object hit by the LiDAR beam. They are used as ground truth data for the point cloud segmentation task.
    \item Vehicle Pose, Speed and Acceleration dataset: computed according to the kinematics of the train trajectory. It contains the 6 DOF (i.e., Degrees Of Freedom) vehicle pose, along with speed and acceleration, allowing generating the ground truth trajectory for the localization algorithms. 
    \item IMU dataset: 6-axes accelerometric and gyroscopic data computed with user-defined IMU models from the ground-truth acceleration and angular velocity data.
\end{itemize}
Figure~\ref{fig:images-example} shows an example of images produced by the RGB and depth cameras, along with the segmented image from the same simulated scene.

\begin{figure}[htbp!]
\centering
\makebox[\columnwidth]{\includegraphics[scale=0.35]{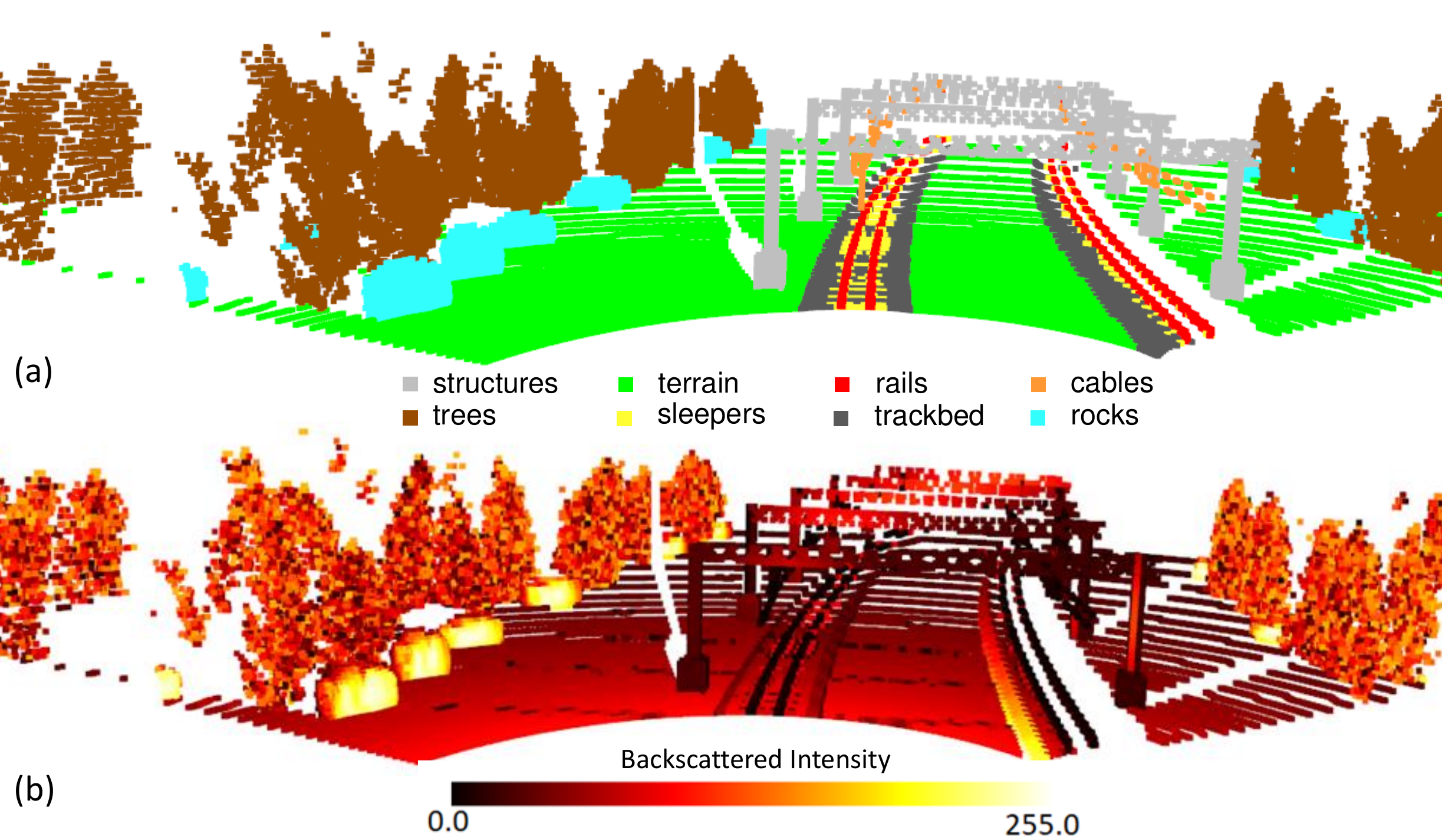}}
\vspace{-5mm}
\caption{(a) Example of a segmented point cloud generated by TrainSim, where each color encodes a different object class; (b) the same point cloud where each color encodes the back-scattered intensity value computed by the LiDAR model.}
\label{fig:example-PC}
\end{figure}

Figure~\ref{fig:example-PC} shows two point clouds of the same scene produced by the virtual LiDAR: in Figure~\ref{fig:example-PC}a each color encodes a different object class, whereas in Figure~\ref{fig:example-PC}b the color encodes the back-scattered intensity.

To summarize, the TrainSim framework aims at providing datasets and ground truth data for the following tasks: Visual Odometry, LiDAR Odometry, Image Segmentation, Point Cloud Segmentation, Image Depth Estimation, and Inertial navigation.

The generation of datasets for tasks like 2D and 3D object detection is in progress and will be part of future work.

The rest of the paper is organized as follows.
Section \ref{s:related} discusses the related works; Section \ref{s:frame} presents TrainSim; Section \ref{s:results} reports some experimental results; and Section \ref{s:conclusions} states the conclusions and future work.

\section{Related Works} 
\label{s:related}

The design of proper datasets for training and testing purposes is crucial for developing and verifying effective perception algorithms.
The tools developed for the automotive domain typically use benchmarks that provide several visual frames captured in different environments, such as the KITTI benchmark~\cite{Geiger2012CVPR} and its semantic segmentation variant~\cite{behley2019semantickitti}, or the Cityscapes dataset~\cite{cordts2015cityscapes}. 
Most of such datasets are focused on urban scenarios, and the vast amount of images required for training is typically obtained by data augmentation~\cite{fang2020augmented}, mixing real and virtual images.
The lack of open datasets in the railway domain represents a severe obstacle to testing and verifying novel algorithms. Zendel et al.~\cite{zendel2019railsem19} pointed out that, excluding the thousands of labeled images taken from street-view or spectator-view, image datasets of railway environments taken directly from the train are nearly nonexistent. 
Many solutions presented in the literature for the railway domain are tested and verified on private datasets that only include a few hundred data samples for camera and LiDAR frames, as declared by the authors~\cite{tschopp2019experimental}, \cite{zendel2019railsem19}, \cite{sahebdivani2020rail}.

Simulators offer the possibility to test perception and control algorithms in a variety of situations that would be hard to reproduce in the real world.
For this reason, several synthetic generation tools have been presented in the last years to overcome the lack of real datasets, as CARLA~\cite{Dosovitskiy17} for automotive simulation and \textit{AirSim}~\cite{airsim2017fsr} for unmanned aerial vehicles (UAV), both based on the Unreal Engine 4 (UE4)~\cite{unrealengine} graphic engine.
Another tool is AutonoVi-Sim~\cite{best2018autonovi}, which supplies LiDAR frames gathered into a virtual world.

Unfortunately, most of the existing simulators have been developed for self-driving cars and drones, and there is a lack of tools for the railway domain that support the integration of the LIDAR, Camera, IMU an GNSS technologies. This work presents TrainSim, a train simulation framework for generating realistic datasets of images, point clouds, and inertial data to test and validate novel algorithms for tasks such as inertial navigation, object detection, and semantic segmentation in the railway domain.
In particular, the camera model is naturally derived from the graphic engine frame, producing RGB, semantic, and depth images directly from the graphic environment of UE4.
On the other hand, the emulation of the LiDAR sensor exploits the ray-casting system of UE4, which allows the detection of objects between two endpoints, making the LiDAR emulation straightforward.
Unlike CARLA and AirSim, the proposed approach also generates the backscattered intensity of the LiDAR sensor by exploiting a simplified version of the Labertian-Beckmann model~\cite{tian2021analysis} that describes how different surfaces reflect light rays.
More details on the images and point clouds generation models are described in Section~\ref{ss:cameramod} and Section~\ref{ss:lidarmod}, respectively.

\section{Simulation framework} 
\label{s:frame}

\begin{figure}[htbp!]
\centering
\makebox[\columnwidth]{\includegraphics[scale=0.45]{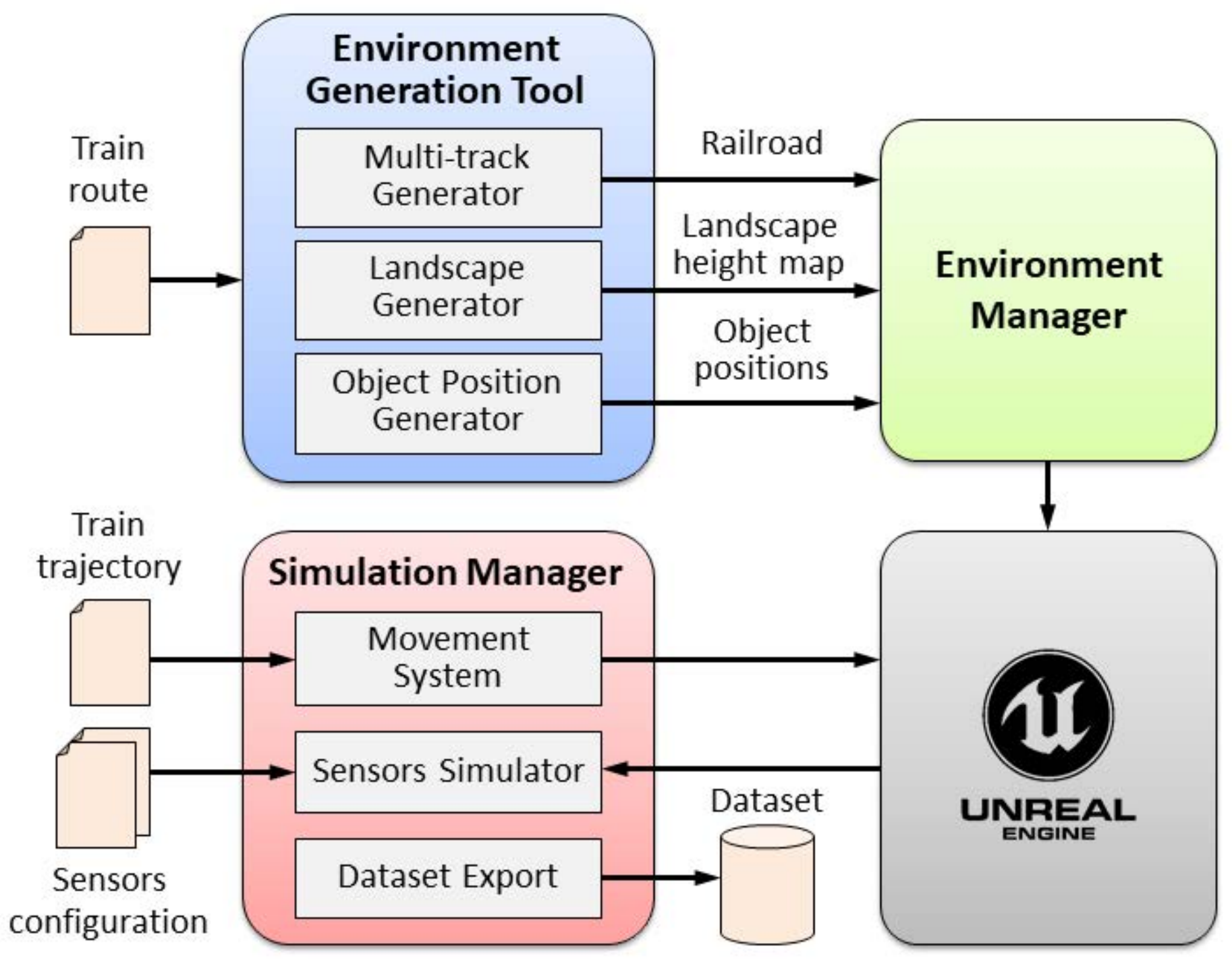}}
\vspace{-5mm}
\caption{Architecture of the TrainSim.}
\label{fig:sim-architecture}
\end{figure}

The architecture of TrainSim is depicted in Figure~\ref{fig:sim-architecture} and is composed of three main modules: the \textit{Environment Generation Tool} (EGT), which manages the creation of the rail-track surrounding environment, the \textit{Environment Manager} (EM), which instantiates the created environment in Unreal Engine, and the \textit{Simulation Manager} (SM), which simulates the train movement, emulates the sensors working principles, and generates the various datasets.

The environment generation is based on the GeoGen project of Matěj Zábský~\cite{geogenstudio}, which is a tool for creating realistic terrains with desired height maps.
The virtual environment is generated starting from the train route specified in a file, which contains set points that are either sampled from a real trajectory or synthetically generated by a separate trajectory generator, described in Section~\ref{ss:trajgen}.

In the following, we refer to a \textit{track} as the physical structure (a pair of rails) where a train can run, and to the \textit{railway environment} as all the ensemble of tracks placed in the environment.
A track is defined as a sequence of 3D waypoints (referred to the track centerline), called \textit{track points} and is divided into \textit{blocks}, where each block identifies a specific type of railway structure, namely a straight line, a curve, a station, a tunnel, or a bridge.
The \textit{main track}, also called the \textit{route}, is the one traveled by the train, whereas the remaining tracks are referred to as \textit{auxiliary}. Then, a \textit{trajectory} refers to a specific train journey on a route (i.e., the sequence of positions, velocities, and accelerations of rear and front bogies of the front vehicle sampled at a given frequency).
The Environment Generation Tool is responsible for managing the creation of the railway environment and consists of three main modules:
\begin{itemize}
    \item The \textit{Multi-track generator} creates a number of \textit{auxiliary tracks} that run parallel to the main track, but can also join it or depart from it with different given rules. It produces a \textit{Railroad} \texttt{.json} file that contains the list of tracks (the main and the auxiliary ones). Each track is divided into blocks (e.g., straight, curve, bridge, etc.) and it is associated with a 3D point sequence and other information needed in the generation of the virtual environment. Refers to Section~\ref{ss:multitrackmod} for further details. 
    \item \textit{Landscape Generator}. It creates the area surrounding the tracks, including the terrain and the mountains. It produces a height-map (i.e., a grid of vertices) in which each vertex is associated with a defined height that derives from the elevation of the tracks. Section~\ref{ss:landscapemod} describes the height-map generation in more detail. 
    \item \textit{Object Position Generator}. It generates random spawn points near the tracks, where different types of objects can be placed (e.g., trees, rocks, buildings). Thanks to its modularity, the placement algorithm can easily be extended to add other types of objects to the scene. In particular, positions are selected taking object size into account to avoid reciprocal overlapping.
\end{itemize}

The outputs of the modules are sent to the Environment Manager, responsible for creating the virtual environment within Unreal Engine by placing the landscapes, the environmental objects, and the railway building structures into the simulated world.
It exploits the 3D object models (meshes, materials and textures) of the \textit{TrainTemplate} plugin~\cite{traintemplate}, which provides high-fidelity models for railways objects, vehicles, stations, tunnels, and bridges.
The meshes for other objects (e.g., trees and rocks) are randomly chosen from a set of different meshes, to diversify the simulated environment both for images and point clouds.
Furthermore, ballast and landscape materials can be randomly drawn at the start of each simulation, avoiding reusing the same texture in each generated dataset.

The Simulation Manager takes as input the train trajectory and a set of files (described in Section~\ref{ss:files}) containing the configuration parameters needed to emulate the working principles of specific sensors.
It tightly interacts with Unreal Engine sending the sequence of train positions and receiving the data produced by the simulated sensors. It includes three main modules:
\begin{itemize}
    \item \textit{Movement System}. It manages the train movement, advancing the train in each point of the specified trajectory.
    \item \textit{Sensors Simulator}. It consists of a set of blocks, each responsible for emulating a specific sensor.
    \item \textit{Dataset Export}. It exports the generated dataset saving it on the disk. The generated datasets can also be transferred in real-time to a ROS Bridge application for visualization or test  purposes by exploiting the ROS communication system.
\end{itemize}

The following sections describe the details of the main architecture components, whereas more details about the remaining modules (e.g., Object Position Generator, Environment Manager) are described in the supplementary material.

\subsection{Input files}
\label{ss:files}

\begin{figure}[htbp!]
    \centering
    \makebox[\columnwidth]{\includegraphics[scale=0.28]{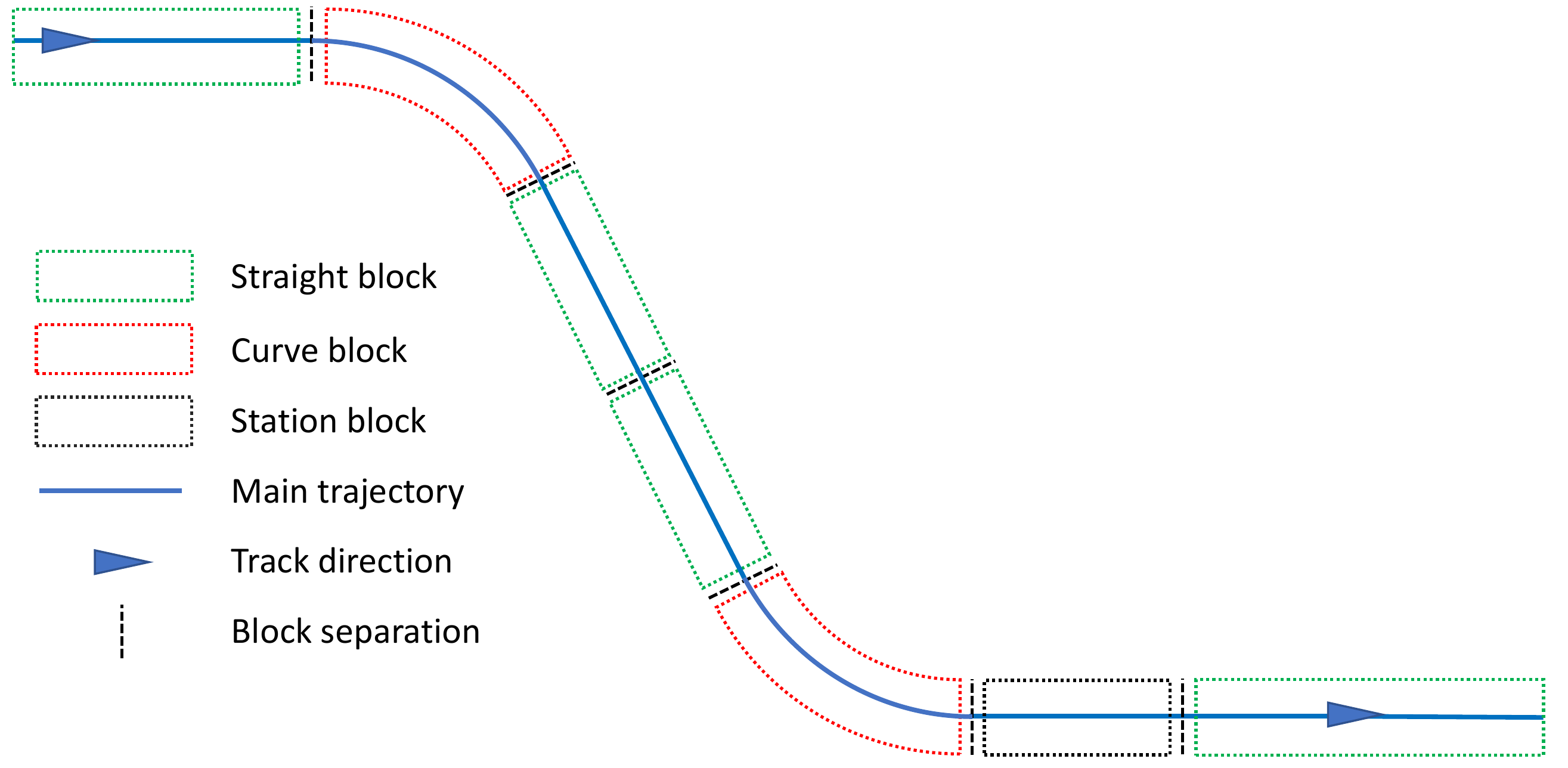}}
    \vspace{-5mm}
    \caption{Example of a railroad section divided into blocks.}
    \label{fig:trackdivision}
\end{figure}

The train route is specified in a file as a sequence of 3D points
$\mathcal{P} = \{P_k \, | \, k = 1, \ldots, N\}$ 
in local north-east-down (NED) coordinates~\cite{groves}.
This sequence is used to generate the corresponding main track, which is divided into construction blocks of different types (e.g., straight, curve, station, bridge, etc.). Each block type has specific characteristics that constrain the construction of the relative way-point sequence and the velocity profile (e.g., curve blocks have a minimum curvature radius and constrain the maximum travelling speed of the train). 
An example of a railroad section divided into blocks is illustrated in Figure~\ref{fig:trackdivision}.

The train trajectory file specifies the position, the velocity, and the acceleration of both front and rear of the vehicle at each timestamp. The trajectory can either be sampled from real IMU and GNSS sensors, or it can be synthetically generated by a proper tool, briefly described in Section~\ref{ss:trajgen}.

Each sensor configuration file provides information on a specific sensor, describing its type, features, parameters, and noise models. These data allow the Sensors Simulator to produce a realistic output by applying the noise models to the data acquired in UE4.

\subsection{Multi-tracks generation} 
\label{ss:multitrackmod}

The Multi-track generator randomly creates a number of additional tracks next to the main track to populate the railway environment. This can be useful to test the performance of track discrimination algorithms.
The user can also decide to duplicate the point sequence of the main track to have double track instead of a single one.
The duplicated track is generated to the right of the main track, since the train hand of drive is on the left at a fixed inter-track distance defined by the user.

\begin{figure}[htbp!]
    \centering
    \makebox[\columnwidth]{\includegraphics[scale=0.28]{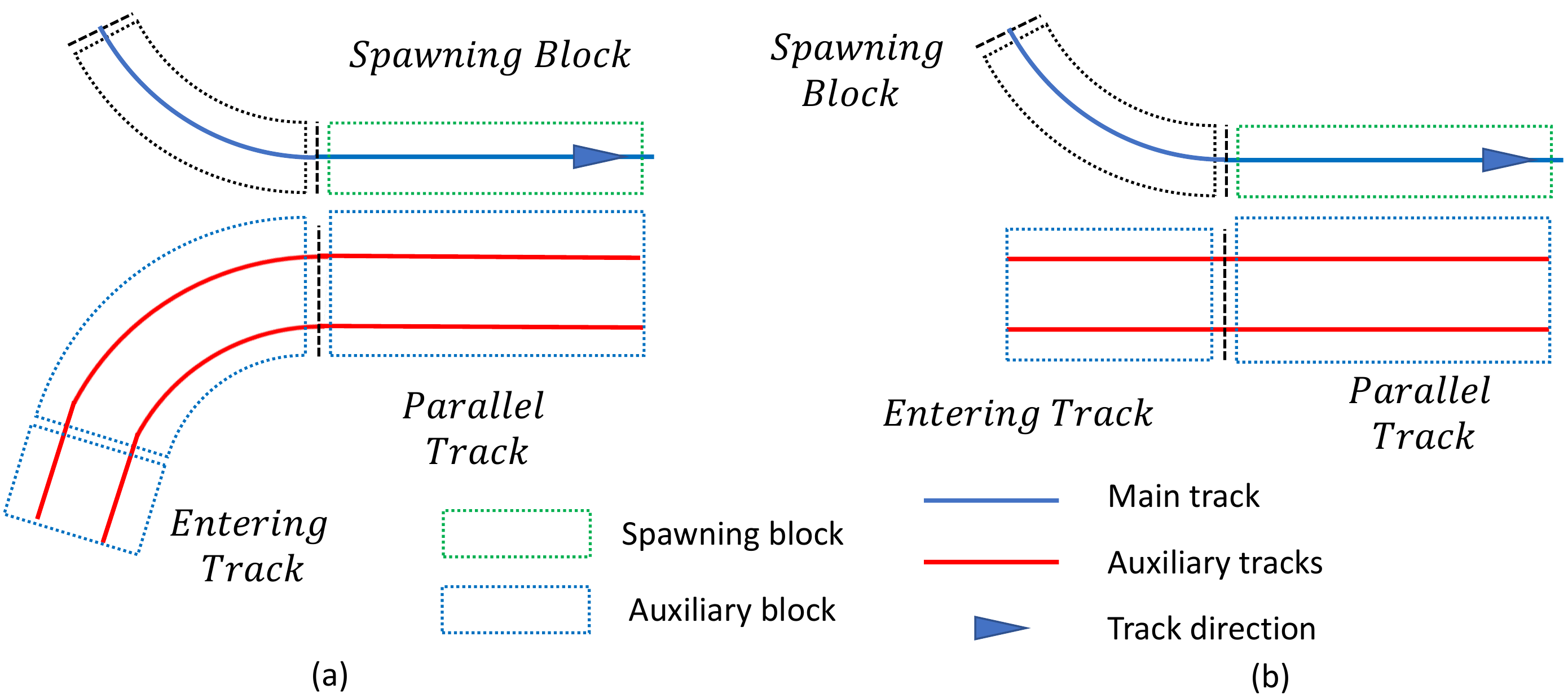}}
    \caption{Entering part examples with 2 parallel tracks, a single line (blue or red) represents a single track): (a) the auxiliary track is generated in correspondences of a straight block, hence it is composed of a straight and a curve blocks; (b) the auxiliary track is generated in correspondence of a curve block, hence it can be composed of a straight block only.}
    \label{fig:enteringtrack}
\end{figure}

To generate additional tracks, the module parses the list of railroad blocks of the main track to decide where to begin or end auxiliary tracks, following the constraints imposed by each track block.
Figure~\ref{fig:trackdivision} shows an example of a railroad blocks division, given as input to the Multi-tracks generator.
An auxiliary track has three parts: an entering part, a parallel part, and an outgoing part.
The entering part is composed of a straight dead-end block, and a curve that joins it to block parallel to the main track.
If the railroad block is a curve, the entering part can be composed only of a straight block, as depicted in Figure~\ref{fig:enteringtrack}.

Some of the building rules are derived from the railway standards~\cite{rfiarmamento,trackdesign}, such as the inter-track distance or the minimum curve radius, whereas others need to be user-defined, like the number of auxiliary tracks.
Once the entering auxiliary block is generated, the iteration proceeds to the next block of the main track, where the parallel parts of the auxiliary track are created at a distance equal to the inter-track distance.

When the auxiliary track ends, the outgoing part is generated from the parallel one, creating a curve block followed by a death-end straight block.
As for the entering part, if the corresponding railroad block is a curve, the outgoing part can be composed of a single straight track line.

The creation of auxiliary tracks follows pseudo-random decisions based on a user-defined probability. 
In this way, the user can manage the auxiliary tracks generation, creating different scenarios from the same train route.

\subsection{Landscape Generation}\label{ss:landscapemod}

Once the auxiliary tracks have been generated, the landscape generation module creates the terrain of the virtual environment, containing the ground, mountains, and valleys.

In UE4, a landscape is defined from a height map, which is a matrix $M$ of vertices, referred to as \textit{main map}, in which each vertex $v_{i,j}$ has its own height value $M_{i,j}$. The main map has a rectangular size that includes all track points.

The final height map is produced by GeoGen~\cite{geogenstudio}, which is a library that allows manipulating height maps using different operations, such as noising, scalar multiplication, and composition.

\begin{figure}[htb]
    \centering
    \makebox[\columnwidth]{\includegraphics[scale=0.5]{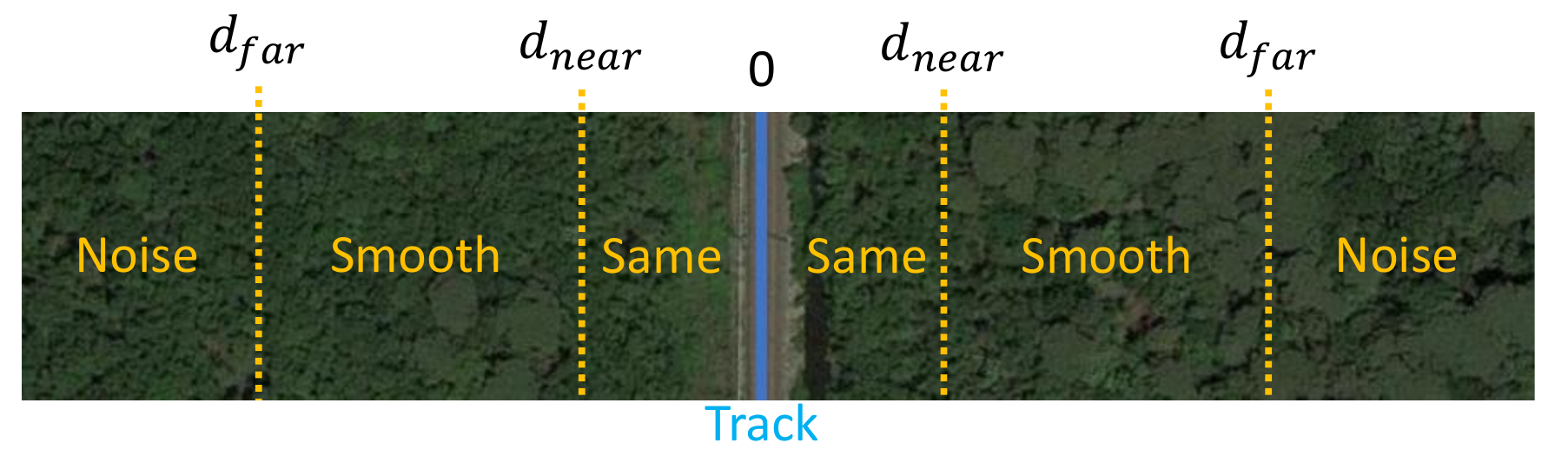}}
    \caption{Example showing how the terrain surrounding the track is partitioned in three areas: the area denoted as \textit{Same} has a height equal to the closest track point; the area denoted as \textit{Noise} has a noisy height; while the area in the middle (\textit{Smooth}) smoothly changes the height between the two values.}
    \label{fig:heightvalues}
\end{figure}

The landscape is partitioned into three different sections with respect to its distance from the outermost tracks, as illustrated in Figure~\ref{fig:heightvalues}.

\begin{figure}[htb!]
    \centering
    \makebox[\columnwidth]{\includegraphics[scale=0.4]{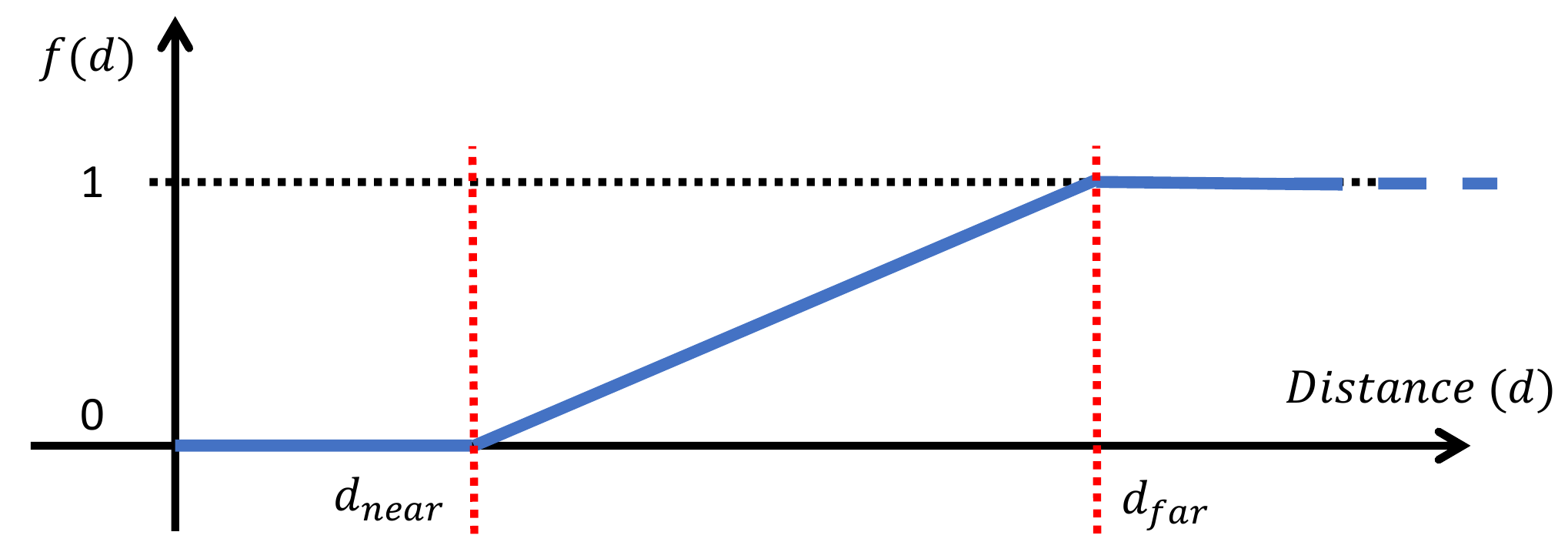}}
    \caption{Distance function used to set the height of the vertices located in the smooth region at a distance from the track between $d_{near}$ and $d_{far}$.}
    \label{fig:distfun}
\end{figure}

In particular, if the minimum distance $d$ of vertex $v_{i,j}$ to the track is less than or equal to $d_{near}$, its height is set equal to the one of the most immediate track point. If the distance to the track is greater than $d_{far}$, the height is sampled from a noise function $N_{i,j}$, using the method proposed in~\cite{geogenstudio}.
Finally, if $d$ is between the two thresholds, the assigned height grows linearly between the two values according to the distance function $f(d)$ illustrated in Figure~\ref{fig:distfun}.

The two distance bounds $d_{near}$ and $d_{far}$ can be set by the user. Please note that $d_{near}$ has a minimum value imposed by the railway construction standards~\cite{rfiarmamento,trackdesign}, namely $1.5$ $m$.
This solution allows easily placing a number of environmental objects (e.g., trees) around the railway structure.

More specifically, given a vertex $v_{i,j}$ in the main map, let $P_n(v_{i,j})$ be the track point with the minimum distance to $v_{i,j}$ and let $d_{i,j}$ such a distance. For the sake of clarity, the point coordinates are expressed in the East-North-Up (ENU) reference system, and $P_n^U$ represents the Up component (i.e., the height).
Then, the height value $M_{i,j}$ associated with vertex $v_{i,j}$ is computed as
\begin{equation}
\label{eq:height}
    M_{i,j} = P_n^U(v_{i,j}) * [1 - f(d_{i,j})] + N_{i,j} * f(d_{i,j}).
\end{equation}
Furthermore, the sub-module generates a valley patch (i.e., a sub-matrix of vertices) that is superimposed to the main map every time a railway bridge is present in the trajectory, so allowing to possibly create a river in that specific position. In station blocks, the value of $d_{near}$ is increased to accommodate buildings and other structures.

\begin{figure}[htb!]
    \centering
    \makebox[\columnwidth]{\includegraphics[scale=0.6]{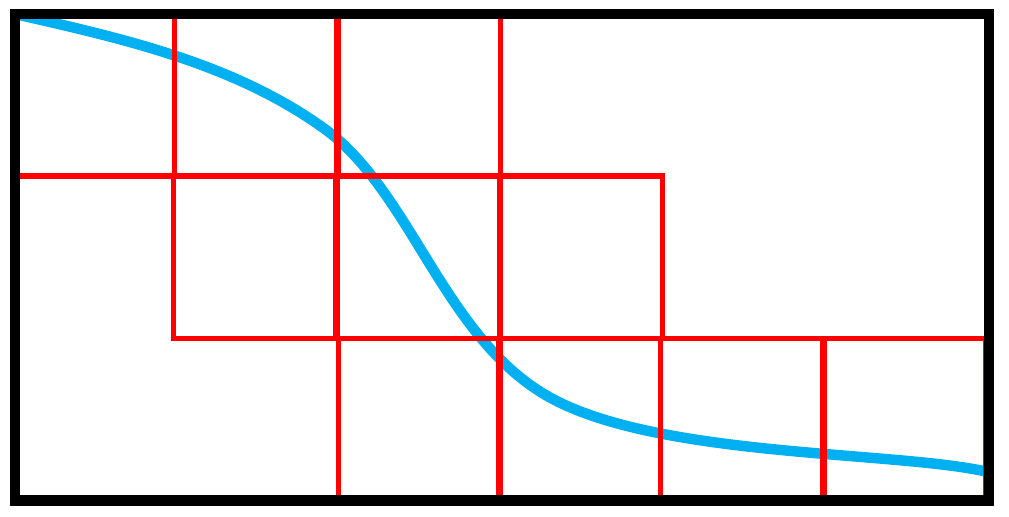}}
    \caption{Example of sub-matrices (red squares) that can be selected based on their distance from the track (blue line).}
    \label{fig:submatrix}
\end{figure}

Once the whole main map is generated, it is divided into sub-maps, so that the user can decide to save only those sub-matrices near the track, in order to save memory.
An example of sub-matrices is depicted in Figure~\ref{fig:submatrix}.

\begin{figure}[htbp!]
    \centering
    \makebox[\columnwidth]{\includegraphics[scale=0.45]{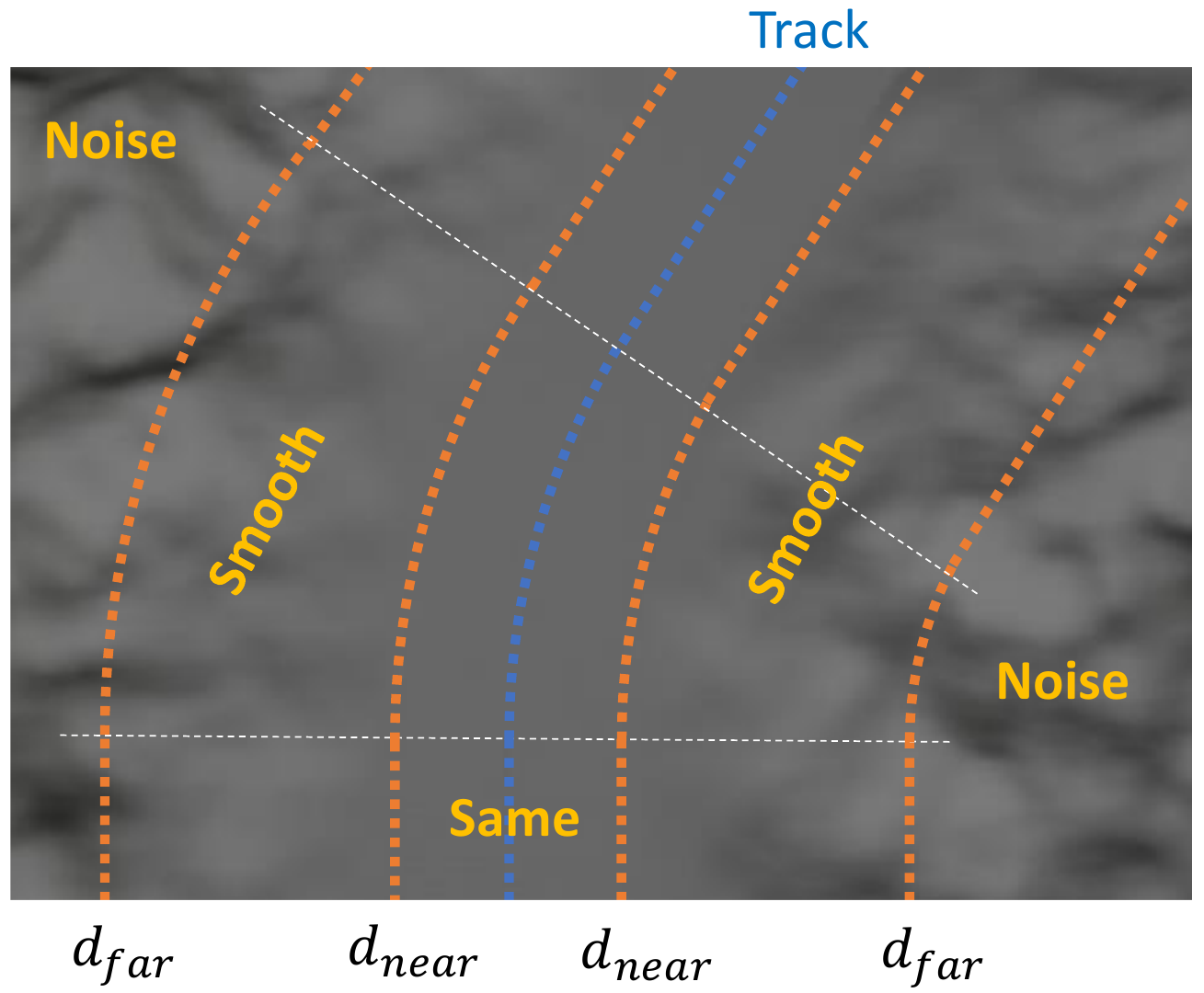}}
    \caption{Example of a UE4 landscape element generated from a height-map.}
    \vspace{-3mm}
    \label{fig:ue4land}
\end{figure}

In the proposed implementation, a sub-matrix has $1009 \times 1009$ vertices.
A scale factor on the E-N axes equal to $100$ is required since the UE4 measurement unit is expressed in $cm$.
With this setting, the distance between two vertices in the horizontal and vertical direction results to be $1$ $m$.
Figure~\ref{fig:ue4land} shows an example of a single landscape element generated with the proposed procedure.

\subsection{Movement System} 
\label{ss:movemod}

The movement system is responsible for updating the position of the train along the route, following the train trajectory specified in the corresponding input file. Each trajectory point includes the position, speed, and acceleration of the front and rear bogies, as well as the corresponding timestamp.

To reproduce the train motion according to the given trajectory, the positions of the bogies have to be computed at each frame by interpolating the position of the two consecutive points in the trajectory that are before and after the tick absolute time. This solution, however, gives rise to two distinct problems:
\begin{enumerate}
    \item The interpolation introduces an error on the virtual position of the train that increases with the train speed (the higher the speed, the higher the distance between trajectory points).
    \item If graphic frames are visualised at a time that is different from the timestamps associated with the trajectory points, then the data produced by virtual IMUs and GNSS receiver are not synchronized with those produced by visual sensors (cameras and LiDARs), hence they are not consistent.
\end{enumerate}

\begin{figure}[htb!]
    \centering
    \makebox[\columnwidth]{\includegraphics[scale=0.35]{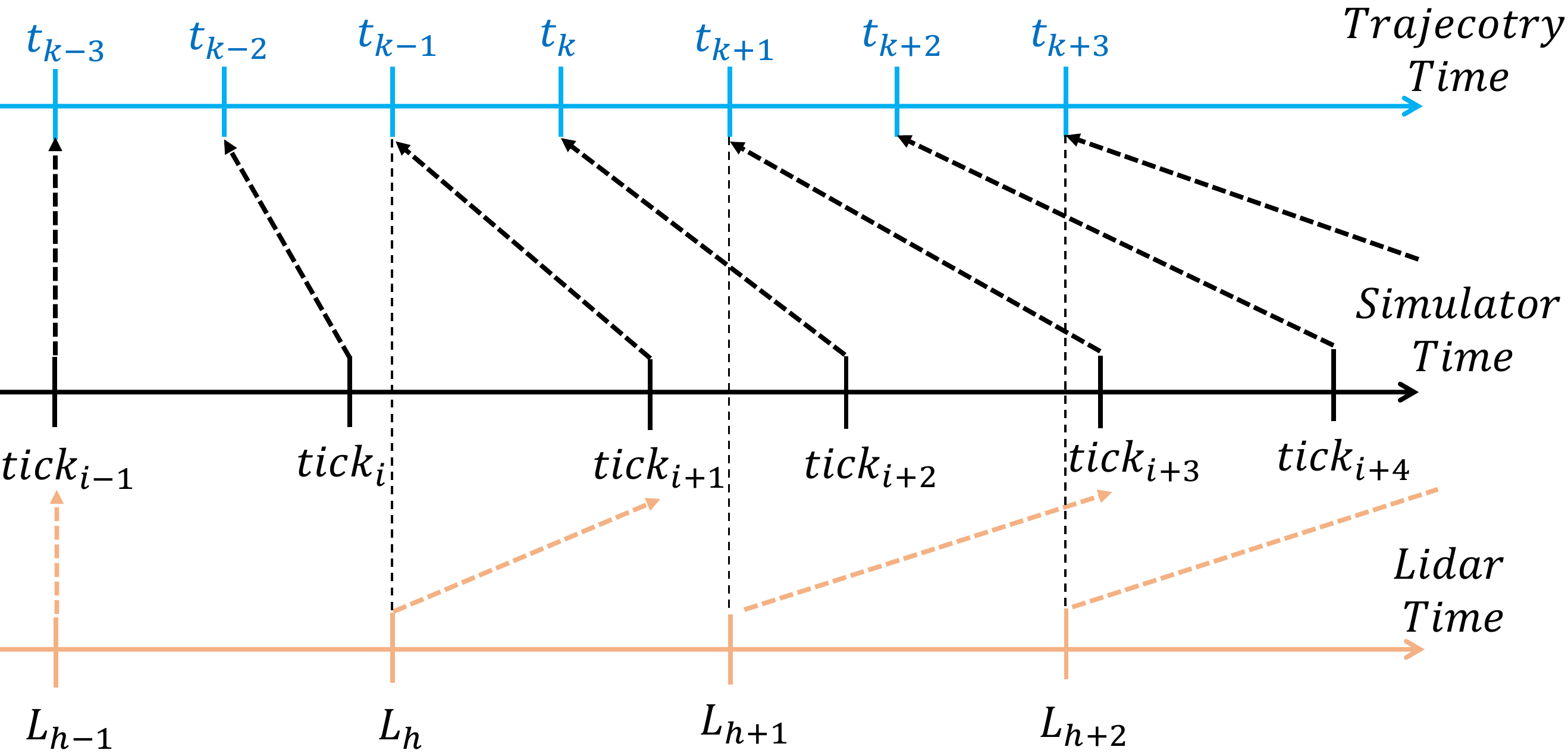}}
    \caption{Timelines corresponding to the train trajectory, the simulator frames, and LiDAR acquisitions. Black dashed arrows show the timestamps associated with the graphic frames, while brown dashed arrows show the graphic frames associated with the LiDAR acquisitions.}
    \label{fig:realtime}
\end{figure}

To address these problems, the \textit{real time} associated with the train trajectory has been decoupled from the \textit{simulated time} at which UE4 produces a visual frame. While the difference between the time stamps of any two consecutive trajectory points is constant and equal to the sampling period $T_S = t_k - t_{k-1}$, the time elapsed between two consecutive frames can vary depending on the machine running the graphic engine.
Hence, each frame produced by the graphic engine must be associated with a trajectory point and its corresponding timestamp, so ignoring the simulation time.
Figure~\ref{fig:realtime} compares the timelines associated with the trajectory, the simulator frames, and the acquisitions from a LiDAR sensor, visualizing the time stamps associated with each frame.

In the example shown in Figure~\ref{fig:realtime}, the LiDAR is acquired with a period that is twice the one used for the trajectory. Black dashed arrows show the timestamps associated with the graphic frames, while brown dashed arrows show the graphic frames associated with the LiDAR acquisitions. In the current implementation, the acquisition period of a visual sensor must be a multiple of the sampling time of the trajectory points.

\subsection{LiDAR Sensor Model} 
\label{ss:lidarmod}

Light Detection And Ranging (LiDAR) sensors are active devices that emit light rays and compute the time needed for the rays to be backscattered to the sensor receivers, or the phase change of the backscattered ray.
If the emitted rays are backscattered, the distance between the LiDAR and the object hit by the ray can be computed from the travelling time and the speed of light, or from the phase difference between the emitted ray and the backscattered one.
Namely, a LiDAR sensor emits several laser beams (or a flash light) and uses a matrix of receivers to create a depth map of the surrounding environment, referred to as a \textit{point cloud}.

Common LiDARs emit a vertical array of laser beams that rotates around the vertical axis to acquire the surrounding scene.
The angular inclination of each laser beam defines the vertical resolution of the sensor, whereas the horizontal angular step at which the rays are emitted defines the horizontal resolution.
The \texttt{json} input file for a LiDAR specifies the horizontal resolution, the horizontal and vertical field of view (\textbf{FOV}), the number of beams (from which it is possible to derive the vertical resolution knowing the vertical FOV), the range of the laser beams, and the frame rate.

The LiDAR working principle is emulated by exploiting the ray tracing system of UE4. In particular, the \texttt{SingleLineTraceByChannel} function of UE4 generates a ray from a starting point to an ending point given as inputs.
If there are objects along the ray, this function returns the closest 3D point in which the ray intersects the first object surface.
It also returns a reference to the object hit, from which it is possible to retrieve other object features stored in the system. Hence, the relative position of the hit point is computed  by subtracting the absolute position of the starting point. 

A single LiDAR frame acquisition is implemented as two nested loops, in which the outer loop iterates over the horizontal angle, and the inner loop iterates along the vertical angles. For each ray, the starting point is the central position of the sensor, while the ending point is the point lying on a beam ray at a distance corresponding to the sensor range.

\begin{figure}[htb!]
    \centering
    \makebox[\columnwidth]{\includegraphics[clip,scale=0.32]{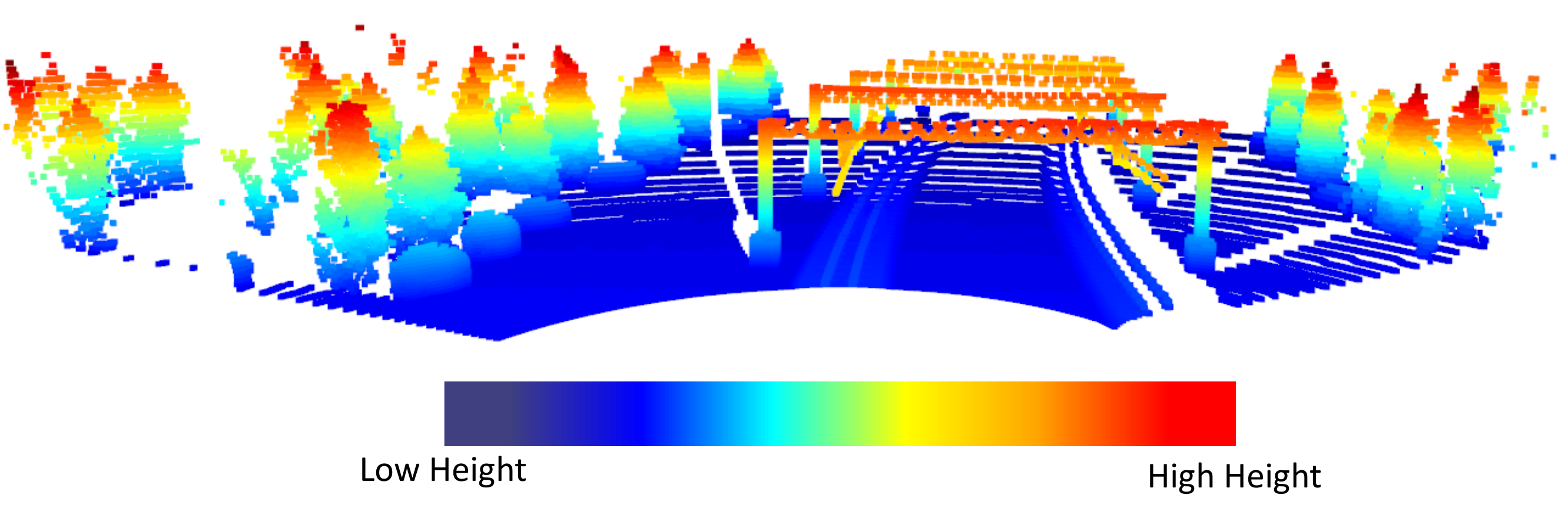}}
    \caption{Example of a point cloud captured from the simulated environment, where each point is colored according to the height value of the point itself.}
    \label{fig:height-pointcloud}
\end{figure}

The sensor is implemented as a UE4 actor component that is positioned on the front vehicle of the train and follows its position at each simulator tick.
A point cloud is produced with the specified period, which must be a multiple of the period at which the trajectory points are generated. For example, if the trajectory period is equal to 10 $ms$ and the LiDAR period is equal to 100 $ms$, a point cloud is generated every ten frames produced by the graphic engine.
Figure~\ref{fig:height-pointcloud} depicts a sample point cloud captured in the simulated environment.

\begin{figure}[htb!]
    \centering
    \makebox[\columnwidth]{\includegraphics[clip,scale=0.6]{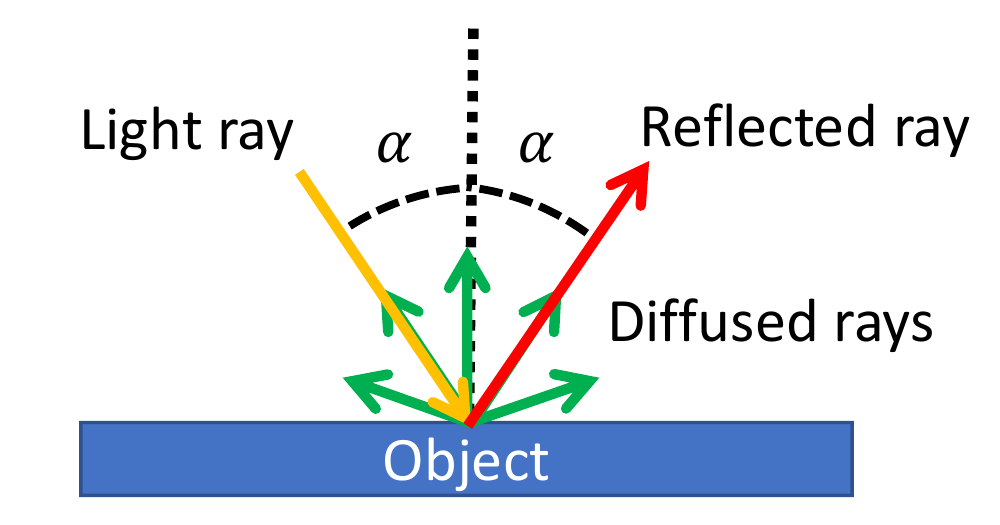}}
    \caption{Object (blue rectangle) response to an emitted light ray (yellow arrow). Part of the light energy is diffused in all directions (green arrows), while an other part (red arrow) is reflected in the opposite direction of the emitted ray.}
    \label{fig:reflmodel}
\end{figure}

Note that, in the real world, if the LiDAR is moving,

the 3D points belonging to a full scan refer to different LiDAR positions, whereas, in the simulated framework, since a graphic frame is a state of the environment frozen in time, all the acquired 3D points refer to the same LiDAR position.
To compensate for such a mis-alignment due to motion, most of LiDAR sensors incorporate an IMU that automatically compensates for the LiDAR movements during a scan. For this reason, the virtual LiDAR implemented in the simulator does not 
take this phenomenon into account. 
To make the data more realistic, the distance associated with each LiDAR beam is perturbed by adding a Gaussian noise with zero mean and given variance.
The object identifier returned by the \texttt{SingleLineTraceByChannel} function is used to assign each point a label, so allowing to create point cloud datasets for semantic segmentation.
Note that each object in the virtual environment is associated with an identifier that specifies the class of the object and the instance (e.g., \texttt{Rock\_0}). 
Thus, an instance segmentation of the point clouds can be obtained as long as each mesh placed in the environment is mapped within the desired set of semantic classes.
Real LiDAR sensors also provide the intensity of each backscattered ray, in terms of light energy, which can be used to discern objects in the environment.
As shown in Figure~\ref{fig:reflmodel}, a fraction of the incident ray is reflected in the opposite direction (red arrow) with an angle equal to the incidence angle, but opposite with respect to the normal to the object surface, whereas another fraction is diffused in all directions (green arrows). The backscattered ray detected by the LiDAR is the diffused ray reflected back in the same direction of the incident ray.

As proposed by Tian et al.~\cite{tian2021analysis}, we used the Labertian-Beckmann model to compute the backscattered intensity as a function of three factors:
\begin{enumerate}
  \item The distance between the sensor and the object;
  \item The incident angle between the emitted ray and the normal to the object surface;
  \item The material of the object.
\end{enumerate}
In particular, each object is associated with a diffusive and reflective coefficient and a maximum incidence angle, over which the object results to be completely reflective.
The model can also be extended to consider other parameters of the environment, as air density and humidity, which, at present, are not taken into account.

To reproduce the backscattering effect in the LiDAR simulation, the incident angle and the material of the object are needed.
The object reference gathered by the \texttt{SingleLineTraceByChannel} function provides the normal to the object surface, which is used to compute the ray incident angle.
Furthermore, a mapping between each object class in the virtual environment and the material parameters needed in the Labertian-Beckmann model is defined and exploited to compute the response of each single object to the LiDAR rays.
In this work, the parameters of different material are taken from the study presented by Tian et al.~\cite{tian2021analysis}. 
A fine tuning in the railway environment would lead to a higher realism of the backscattered intensity of the simulated point cloud.
Figure~\ref{fig:example-PC}a shows an example of a point cloud where each point is colored with the class of the corresponding object, and Figure~\ref{fig:example-PC}b shows the same point cloud where each color encodes the backscattered intensity value normalized as an integer in the range $[0,255]$.

\subsection{Camera Sensor Model}
\label{ss:cameramod}

Unreal Engine 4 allows the user to create cameras to gather images of the virtual environment from user defined-locations.
In particular, a camera can be placed in front-top of the locomotive to capture an image at each tick of the graphic engine.
As for the LiDAR sensor, the camera capture period can be specified as a multiple of the one used for the trajectory. The user can also define a number of parameters of the camera, such as shutter speed, aperture, ISO, and resolution.

Additionally, UE4 allows generating a depth image of the scene, where each pixel value encodes the distance between the camera and the object represented by the pixel. 
The distance is normalized into a range $[0, depth_{max}]$, where all the values above $depth_{max}$ are cut off and set equal to $depth_{max}$. The value of $depth_{max}$ is set to $100$ $m$ by default and can be redefined by the user.

UE4 also allows defining post-processing routines that exploit custom stencils to create a segmented version of an RGB image.
Each type of object placed in the virtual environment is assigned a specific custom stencil value used to distinguish each object in a segmented image.
Figure~\ref{fig:images-example} shows an RGB image (left image) with the corresponding depth image (middle image) taken from the same scene, and along with the corresponding segmented one (right image), where each object class is identified by a different color.

\begin{figure}[htb!]
\centering
\makebox[\columnwidth]{\includegraphics[scale=0.34]{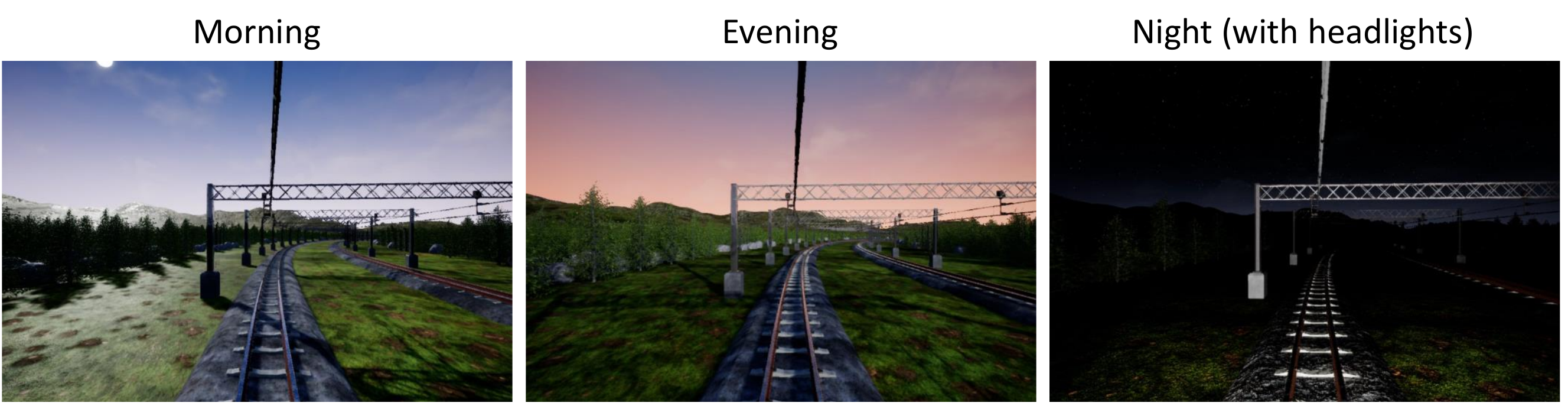}}
\vspace{-5mm}
\caption{Examples of RGB images generated by TrainSim at different daytimes: left, morning; middle, evening; right, night.}
\label{fig:image-daytime}
\end{figure}

At last, TrainSim allows defining different ambient aspects and weather conditions used to test visual based-algorithms in a wide range of operating conditions.
In particular, it is possible to define:
\begin{itemize}
    \item the Sun position, for creating images with different shadows and light intensity. The framework defines three different time slots, morning, evening, and night, as shown in Figure~\ref{fig:image-daytime};
    \item the fog, with a desired intensity, by inserting the \textit{ExponentialHeightFog} UE4 actor in the environment.
\end{itemize}

\subsection{IMU model} 
\label{ss:imumodel}

The proposed simulation framework includes a model of a 9-axis inertial measurement unit (IMU) with accelerometers, gyroscopes, and magnetometers, for estimating the current position, velocity, and orientation of the train by means of inertial navigation algorithms~\cite{groves}. To reproduce realistic data with high fidelity, the IMU model allows the user to specify noise properties, calibrated bias, and other parameters that affect the quality of the measures.

The simulated measured quantity $\widetilde{a}$ in the IMU reference frame is computed from the ground-truth quantity $a$ in the NED frame by means of the accelerometer model $\mathcal{A}$ to obtain $\widetilde{a} = \mathcal{A}(a, \theta)$, where $\theta$ is the orientation of the IMU, necessary to return readings in the IMU frame.

Function $\mathcal{A}$ depends on the following factors:
\begin{enumerate}
    \item The gravitational acceleration $g$, added along the Down component of $a$ and converted to the IMU frame using the rotation matrix $C_{NED}^{IMU}(\theta)$ computed from the orientation $\theta$.
    \item The misalignment matrix $Mis$ (due to geometrical imperfections of the orientation of the individual accelerometer axes) and a constant calibrated bias $\epsilon$, used to alter the ground-truth acceleration.
    \item A drift term $\delta$, which depends on noise parameters, such as Bias Instability, Noise Density, Random Walk, and environmental causes, as temperature-induced bias.
    \item A quantization factor Q, used to replicate the resolution of the sensor.
\end{enumerate}
The resulting function for the accelerometer is then:
\begin{equation}
    \widetilde{a}=\mathcal{A}(a, \theta) = \text{Q}\left(Mis ~ C_{NED}^{IMU}(\theta)(a+g) + \epsilon + \delta\right).
\end{equation}

Similar formulations are used for simulating the outputs of gyroscopes and magnetometers. Changing the noise, bias, resolution, or any other parameter in the datasheet of a specific sensor allows simulating different IMUs.

\subsection{Trajectory Generator}\label{ss:trajgen}

This tool generates pseudo-random train routes and related journey trajectories to be used as input files for the simulator.

As a first step, the route is generated as a sequence of curves and straight blocks that follow the constraints imposed by the construction standards. The output is a sequence of spatially evenly distributed points $\mathcal{P} = \{P_k \, | \, k = 1, \ldots, N\}$.
Then, among the straight blocks, some of them are randomly selected as bridges, tunnels, and stations, using constraints and probability provided by the user.
It is possible to interpolate the position between points by fitting three smoothing splines~\cite{gu2013smoothing} on such points $\mathcal{P}$, one for each axis (north, east, down).
This approach allows importing the set $\mathcal{P}$ from the digital map of an actual route.

The second step defines the maximum train velocity on each block of the track, e.g., inside tunnels, on bridges, within stations, and in each curve as a function of its curvature.
The details on the generation of such synthetic routes and velocity profiles are omitted for space limitations, also considering that the algorithm could be extended to better adapt to the final dataset produced by the simulator.

Then, the tool exploits the kinematic model and the control law of the train, combined with the geometry of the route and the maximum velocity profile, to yield the final trajectory of the front and rear bogies of the vehicle (position, velocity, acceleration, and orientation) with the desired sampling time $T_S$.
Note that the orientation at each sampling instant is computed only using the geometry of the track.
In fact, since the motion of the train is heavily constrained by the track, the only allowed orientation of a vehicle is the orientation of the track itself. Hence, the yaw, pitch, and roll angles are computed from the line tangent to the track in the current bogie position. At the same time, the angular velocity is obtained by kinematics as a function of the orientation and orientation rates on the three axes.

The trajectory of a generic point on a vehicle can be computed from the trajectories of its front and rear bogies.

\subsection{Dataset Export} 
\label{ss:dataexportmod}

The generated dataset can be transmitted for online usage or stored to be employed offline.

The proposed online method allows the user to directly connect the UE4 simulator to a ROS network, creating a sensor node that exposes the frame data right after the acquisition, providing a simulation system that can be tested and evaluated online.  
The ROSIntegration plugin~\cite{rosintegration} for UE4 is used to create distinct topics for images, point clouds, and inertial data that are transmitted through a TCP connection to a ROS~\cite{ros} bridge node.

Datasets are saved on the disk with the same data format used by other urban open datasets that can be found in the scientific community, such as the KITTI dataset~\cite{KITTI}.
In this way, most of the automotive algorithms can be tested on the saved train dataset to evaluate their performance in a railway environment without additional pre-processing.
\section{Experimental Results}
\label{s:results}

This section presents some experimental results aimed at testing the realism of the the simulated datasets.
Section~\ref{ss:lidarWPA} compares a real point cloud gathered in a static environment with a point cloud generated by TrainSim on a similar static scene re-created on the graphic engine.
Section~\ref{ss:lidarOdomA} compares the performance of a state-of-the-art LiDAR odometry algorithm on a sequence LiDAR frames generated by TrainSim and taken from the KITTI dataset.
Finally, Section~\ref{ss:imageSemA} compares the results of an image semantic segmentation algorithm applied to both the TrainSim generated data and the RailSem19~\cite{zendel2019railsem19} dataset.

\subsection{LiDAR Working Principle Analysis}\label{ss:lidarWPA}

\begin{figure}[htb!]
\centering
\makebox[\columnwidth]{\includegraphics[scale=0.28]{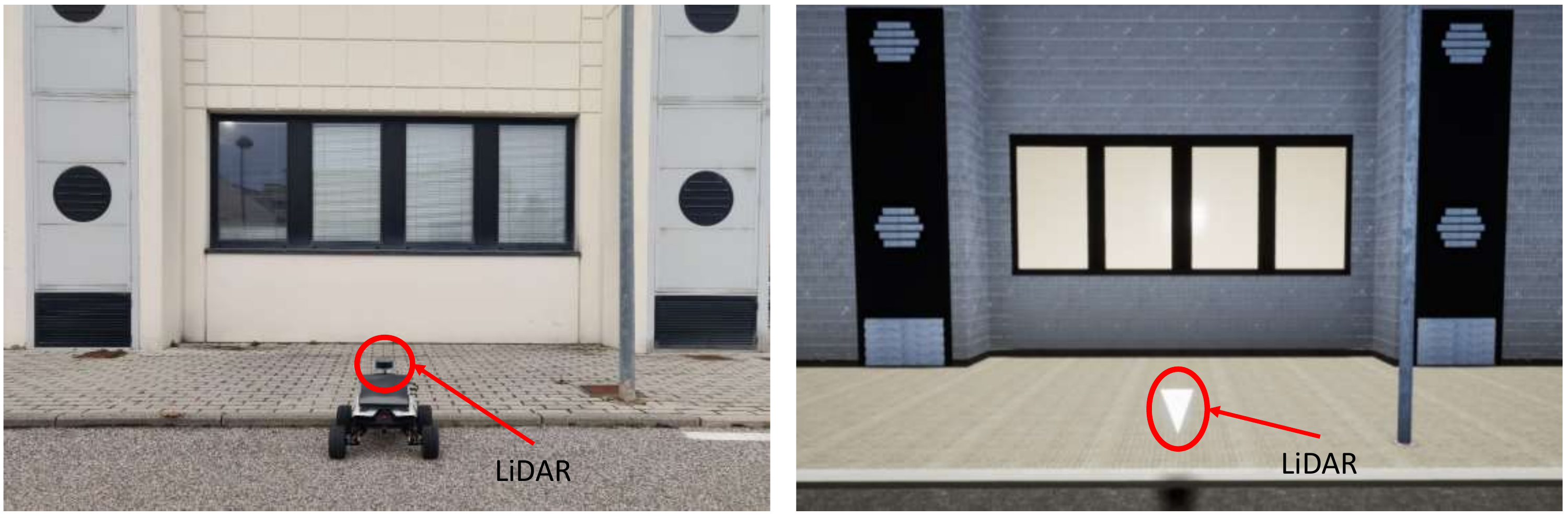}}
\vspace{-5mm}
\caption{Real-world reference static scene (left) and similar scene re-created in the simulation framework (left).}
\label{fig:courtyard}
\end{figure}

This section aims at evaluating the emulation of the working principles of a LiDAR sensor in TrainSim, considering both distance and backscattered intensity measurements. Real point clouds were acquired using a Scout Mini Robot\footnote{https://global.agilex.ai/products/scout-mini} equipped with a Velodyne VLP-16\footnote{https://velodynelidar.com/products/puck/} LiDAR sensor. As illustrated in Figure~\ref{fig:courtyard}, the robot was positioned on an courtyard in front of a wall (Figure~\ref{fig:courtyard}, left) and a similar scenario has been re-created in TrainSim (Figure~\ref{fig:courtyard}, right).

The VLP-16 is a $360^\circ$ rotating LiDAR with 16 vertically aligned laser beams covering a vertical Field Of View (FoV) of $30^\circ$, vertical resolution of $2^\circ$, and horizontal resolution of $0.2^\circ$ for the default rotation speed.
In the following, $\theta$ denotes the yaw orientation angle, while $\phi_i$ and $\rho_i$ denote the vertical angle displacement and the distance reading of the $i^{th}$ beam, respectively.

\begin{figure}[htb!]
\centering
\makebox[\columnwidth]{\includegraphics[scale=0.4]{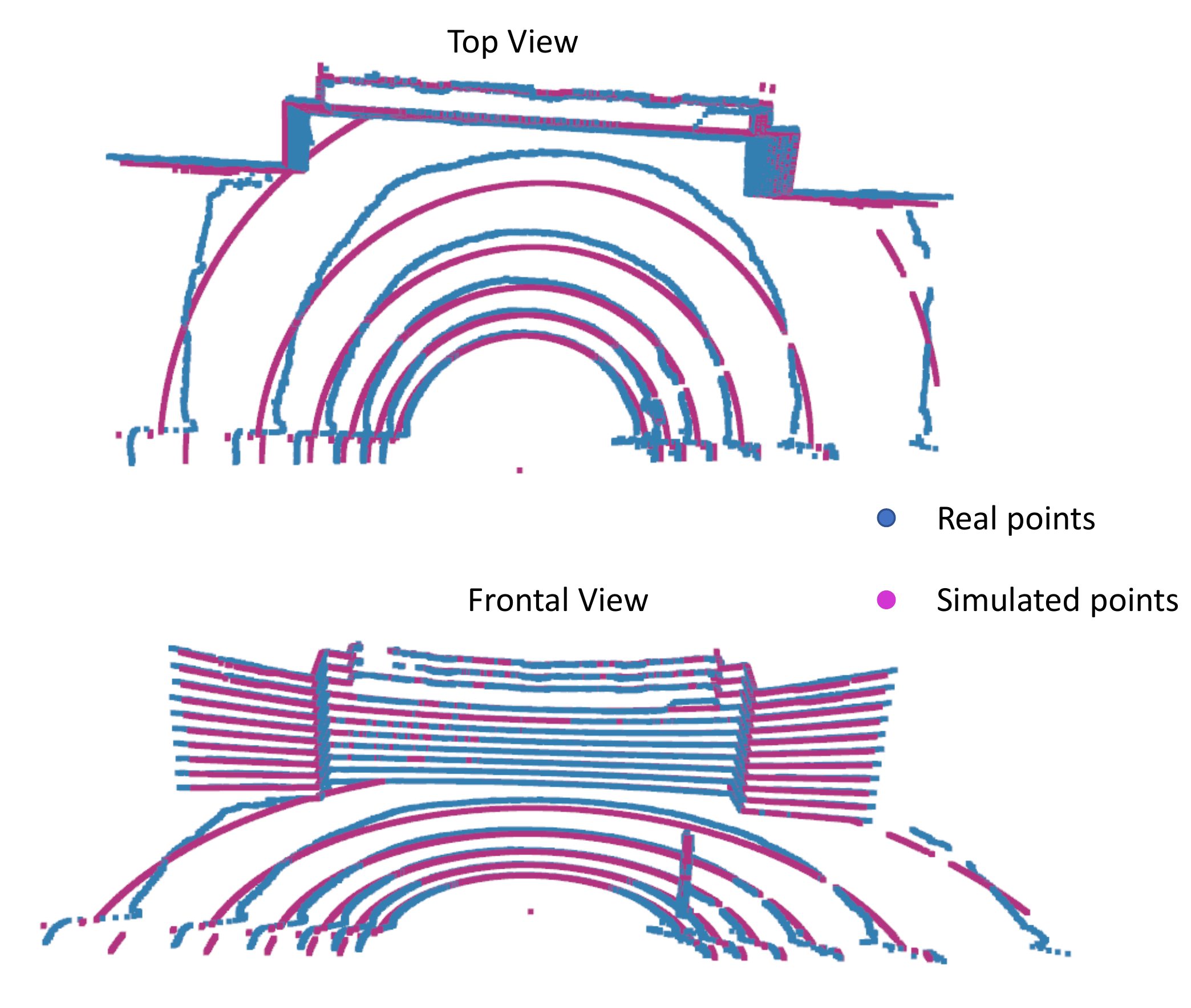}}
\vspace{-5mm}
\caption{Top and frontal view of the real and simulated point clouds.}
\label{fig:both_pointclouds}
\end{figure}

The Root Mean Square Error (RMSE) between the real and the simulated cloud points was computed to evaluate the realism of the LiDAR simulation, as done in \cite{gusmao2020development}.
To be more consistent with some restrictions in the reconstructed environment, the point clouds have been cropped to reduce the horizontal FoV, setting  $\theta \in [-\frac{\pi}{2}, \frac{\pi}{2}]$.
It is worth noting that the simulated virtual scene has been re-constructed manually, introducing position errors due to measurement errors and shape misalignment imprecisions, which increased the resulting RMSE.
Figure~\ref{fig:both_pointclouds} illustrates the top and frontal view of the two point clouds (real data are presented in blue and simulated ones in purple). Note that the largest misalignment between points is due to the irregular shape of the real sidewalk, which is simply represented by a flat surface in the simulation.
Excluding the sidewalk points from the comparison, the RMSE resulted of $0.035$ $m$, which is in accordance with the precision of the VLP-16 LiDAR.

The presented results were obtained with a simulated point cloud without considering the noise, since the datasheet of the LiDAR device reports the  precision only for the distance $\rho$ and not for the ray angular displacements.
Adding to the measured distance $\rho$ a Gaussian white noise comparable with the VLP-16 precision (i.e., zero mean and variance $0.015$) did not significantly change the RMSE, which resulted of $0.081$ $m$ considering the whole point cloud and $0.04$ $m$ excluding the points belonging to the sidewalk.

\begin{figure}[htb!]
\centering
\makebox[\columnwidth]{\includegraphics[scale=0.30]{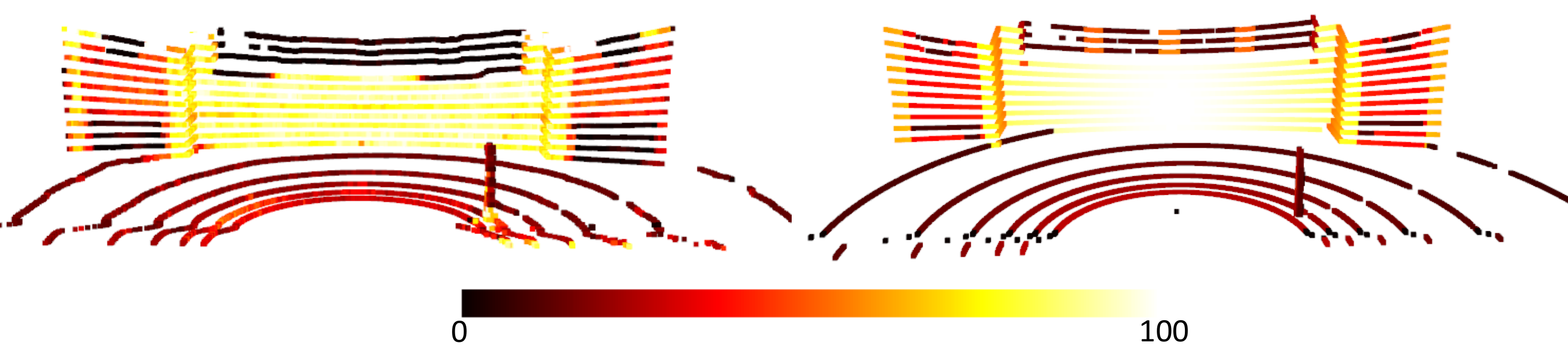}}
\vspace{-5mm}
\caption{Frontal view of the real (right) and simulated (left) point clouds. The color associated with each point encodes the backscattered intensity values.}
\label{fig:intensity_compare}
\end{figure}

Concerning the backscattered intensity, there is not a standard way to process data, thus LiDAR manufacturers use different methods to compensate the measurements with respect to various parameters, as distance and incidence angle (see Section\ref{ss:lidarmod}). Such compensation methods are frequently unknown, making it challenging to precisely reproduce the output backscattered intensity values for a specific LiDAR device.
In particular, VLP-16 divides the intensity values into two subranges: values in $[0, 100]$ map diffuse reflectors with a reflectance in the range $0 - 100\%$, while values in $[101, 255]$ represent retro reflectors with an ideal reflection. Unfortunately, the calibration mechanism is not precisely described in the VLP-16 User Manual, restricting the possibility of reproducing an exact representation of the backscattered intensity in TrainSim.
For this reason, a qualitative comparison between the intensity values is presented, showing the distribution of the intensities with respect to incidence angles, distances, object materials, roughness, and reflectance. The model to account for the last three parameters has been taken from~\cite{tian2021analysis}.
In the reference scene, retro-reflected elements are not present, and the intensity values are scaled in the $[0,100]$ range to match the VLP-16 specifications.
Figure~\ref{fig:intensity_compare} shows the point clouds gathered from the reference scene and the simulated one, showing three different effects that can be underlined:
\begin{itemize}
    \item the backscattered values of the sidewalk in front of the LiDAR decrease by increasing the angle of incidence;
    \item the metal pole has high-intensity values in the frontal part of the pole, rapidly decreasing on the pole boundaries;
    \item the effect of the angle of incidence on concrete material such as the wall is lower than the effect on metallic or plastic material.
\end{itemize}

\subsection{LiDAR Odometry Analysis}
\label{ss:lidarOdomA}

This experiment compares the results obtained with simulated point clouds against real ones on an odometry task. Due to the lack of public point cloud datasets
acquired from a train, the KITTI~\cite{KITTI} urban automotive dataset was select for comparison.

The purpose of LiDAR odometry is to predict the motion of the LiDAR sensor from consecutive LiDAR frames. The ego-motion estimation is done by iteratively computing the homogeneous transformation matrix $T_k$
between two consecutive frames $F_k$ and $F_{k+1}$ that maximize the alignment between the two frames.
Formally, the transformation matrix is defined as $T_k=
\begin{bmatrix}
    R_k & t_k \\
    \bar{0} & 1
\end{bmatrix}$, where $R_k$ is a rotation matrix, $t_k$ is a translation vector, and $\bar{0}$ is a vector of zeros. The best alignment can be defined as an optimization process aimed at minimizing the following distance function:
\begin{equation}
\begin{aligned}
    d_m(T_k) &= \sum_{i=1}^{N_k}{((R_k\cdot p_i + t_k - q_i) }, \\
\end{aligned}
\label{eq:loam}
\end{equation}
where  $N_k$ is the number of points in frame $F_k$, $p_i \in F_k$ is a point in frame $F_k$, and $q_i \in F_{k+1}$ is the point closest to $p_i$ after applying transformation $T_k$ to $F_k$.
From the estimated transformation $T_{k+1}$, computed at time $k+1$, it is possible to predict the ego-motion of the LiDAR sensor in terms of orientation $R_k$ and translation $t_k$.

In this work, the LiDAR Odometry And Mapping (LOAM)~\cite{loam} algorithm was used for the odometry task. 
In particular, the LOAM algorithm is divided into two consecutive modules: (i) an odometry algorithm that is computed at a high frequency with low precision, and (ii) a mapping algorithm that is executed at a lower frequency but with a higher accuracy.
By default, the odometry algorithm  extracts 24 features, whereas the mapping algorithm extracts 240 features to have higher precision, with a ratio of 1:2
between corner and planar features.
Both algorithms extract a fixed number of corner and planar features from frame $F_{k+1}$, find and matches the same features in the frame $F_k$, and iteratively minimize the distance presented in Equation~\ref{eq:loam} to compute the best alignment transformation $T_{k+1}$.

To distinguish between planar and corner features, a feature factor $c$ is computed~\cite{loam} for each point, where a low $c$ value indicates a planar feature, whereas a high $c$ value indicates a corner feature. Then, features are ordered based on the $c$ values and $N$ corner points are selected taking the highest $c$ values, and $2N$ planar points are selected taking the lowest $c$ values, where $N$ is a user-defined parameters (set to 8 for the odometry step and to 80 for the mapping step).

The estimation error of the LOAM depends on the quality of the extracted features: environments containing repetitive features, such as tunnels or highways (hard to be detected in different frames), or with a low number of peculiar features leads to higher estimation errors.
Moreover, since train and car motion mostly evolve in the X-Y plane, with low variations of the Z values, the Z evolution is not observable in such environments, unless the terrain presents substantial Z variations during motion. 

In this experiment, three different sequences of the KITTI dataset have been chosen. The sequence with the identifier 00 is completely gathered in a urban environment with abundant high-quality features.
The second, with identifier 01, is gathered on a highway, which has low-quality features. The third sequence, identified as 09, is divided into two parts: the first is collected in a leaning street with a lot of vegetation on the sides, and the second is gathered in a urban environment.
The comparison is made against three different sequences generated by TrainSim, where the environment is composed of vegetation, railway structures (e.g., poles, electrified structures, and rails), fences, and stations.

Three different metrics have been chosen to evaluate the LOAM algorithm on the selected sequences: the estimation error of the translation along the $X$ and the $Y$ axis on the single transformations between each two consecutive LiDAR frames (\textbf{TEX} and \textbf{TEY}), and the cumulative position error in the $X$-$Y$ plane computed over the traveled distance (\textbf{EOD}).
The translation over $Z$ was not taken into account due to the low variability of the $Z$ coordinates, whereas the orientation estimations were not reported because the estimation error resulted to be below $1^{\circ}$.
Table~\ref{t:loam_results} shows the results of the LOAM algorithm applied to the six sequences.

\begin{table}[]
\setlength\tabcolsep{3.7pt}
\begin{tabular}{l|rr|rr|rr}
\multirow{2}{*}{Sequence} & \multicolumn{2}{c|}{TEX}                        & \multicolumn{2}{c|}{TEY}                        & \multicolumn{2}{c}{EOD}                        \\
                  & \multicolumn{1}{c}{$\mu\pm\sigma$} & \multicolumn{1}{c|}{Max} & \multicolumn{1}{c}{$\mu\pm\sigma$} & \multicolumn{1}{c|}{Max} & \multicolumn{1}{c}{$\mu\pm\sigma$} & \multicolumn{1}{c}{Max} \\ \hline
Sim 1 & 0.06{\footnotesize $\pm$0.04} & 0.18 & 0.08{\footnotesize $\pm$0.05} & 0.17 & 0.25{\footnotesize $\pm$0.15\%} & 1.45\% \\
Sim 2 & 0.34{\footnotesize $\pm$0.18} & 0.69 & 0.54{\footnotesize $\pm$0.28} & 1.18 & 2.47{\footnotesize $\pm$2.06\%} & 22.80\% \\
Sim 3 & 0.15{\footnotesize $\pm$0.10} & 0.38 & 0.09{\footnotesize $\pm$0.15} & 0.89 & 0.83{\footnotesize $\pm$1.26\%} & 7.64\% \\ \hline
Kitti 00 & 0.14{\footnotesize $\pm$0.18} & 1.11 & 0.17{\footnotesize $\pm$0.24} & 1.40 & 0.44{\footnotesize $\pm$0.80\%} & 13.2\% \\
Kitti 01 & 2.17{\footnotesize $\pm$0.71} & 3.57 & 1.04{\footnotesize $\pm$0.51} & 2.76 & 11.40{\footnotesize $\pm$3.68\%} & 45.81\% \\
Kitti 09 & 0.26{\footnotesize $\pm$0.19} & 0.70 & 0.34{\footnotesize $\pm$0.2} & 0.86 & 1.65{\footnotesize $\pm$0.86\%} & 7.22\% \\
\end{tabular}
\caption{\small{Localization results of the LOAM algorithm~\cite{loam} applied to three sequences generated with TrainSim and three similar sequences from the KITTI dataset~\cite{KITTI}. The values indicate the mean $\pm$ the variance and the maximum error value of the translation error along the X axis (TEX), the translation error along the Y axis (TEY) computed in a single estimated transformation between two consecutive LiDAR frames expressed in meters, and the overall error over distance (EOD) in percentage.}}
\label{t:loam_results}
\end{table}

The results indicate that the estimation error obtained on the simulated environment is comparable with the one over the KITTI sequences. In particular, the first simulated sequence is surrounded by vegetation and buildings, whereas the second and the third sequences present some fence stripes that introduce repetitiveness in the distribution of the features; in particular, the second sequence contains some heathland that is comparable with the highway environment.
The same trend can be seen in the KITTI sequence, where the highest error occurs in the KITTI 01 sequence gathered in the highway, while the lowest error is achieved in the KITTI 00, which is entirely acquired in a urban environment.

\subsection{Image Segmentation Analysis}
\label{ss:imageSemA}

In several works in the autonomous driving domain (e.g., \cite{gta_da, SS_DA, UDA, surveyDA}), synthetic scenarios are used with \textit{domain adaptation} (DA) techniques to improve the accuracy of a neural model whenever there is a scarce availability of real-world annotated samples, which is particularly true for the railway domain.
In particular, during training, such techniques help to select learnable features from synthetic images that enhance the model outcome in real-world testing scenarios. Therefore, this section presents an experiment aimed at evaluating the improvement obtained on a neural model when augmenting the training set with synthetic images generated by TrainSim.

To do that, we evaluated the performance of a neural network on a real-world test set by comparing two different training modes: \textit{semi-supervised} (SS) \cite{surveySS} and \textit{semi-supervised with domain adaptation} (SSDA) \cite{surveyDA}.
More specifically, in SS mode, the neural model is trained using only real-world images, following supervised and unsupervised paradigms for labeled and unlabeled samples, respectively. In SSDA mode, instead, the model is trained using the same paradigms for real-world samples, but the training set is augmented with labeled synthetic images.

In the experiment presented here, SSDA was performed via a discriminator approach \cite{SS_DA} and, for consistency, the SS mode was also implemented by a discriminator approach \cite{UDA} using the real-world annotated samples as the source dataset.

Real-world images were taken from the RailSem dataset \cite{railsem}, containing more than 8000 annotated samples collected from both railway and urban scenarios. In our tests, 6000 samples were used for the training set: $6000-k$ with annotations and $k$ without annotations, setting $k=10$ and $k=20$ to observe the difference in performance. Other 2000 samples were used for the real-world test set.

\begin{figure}[ht]
     \centering
     \begin{subfigure}{0.98\columnwidth}
         \centering
         \includegraphics[width=\textwidth]{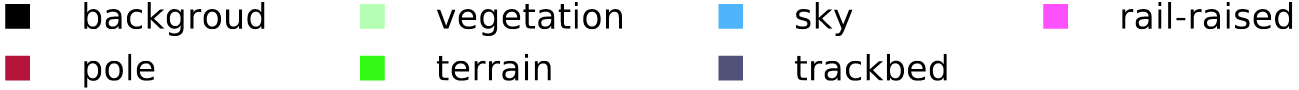}
     \end{subfigure}
     \begin{subfigure}{0.32\columnwidth}
         \centering
         \includegraphics[width=\textwidth]{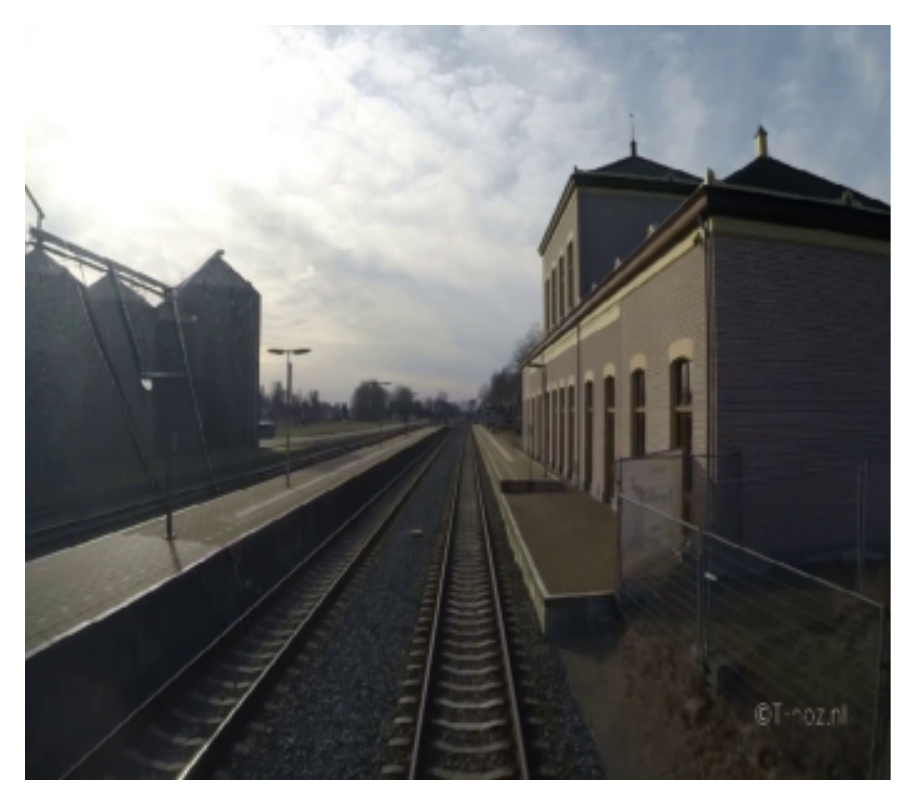}
     \end{subfigure}
     \begin{subfigure}{0.32\columnwidth}
         \centering
         \includegraphics[width=\textwidth]{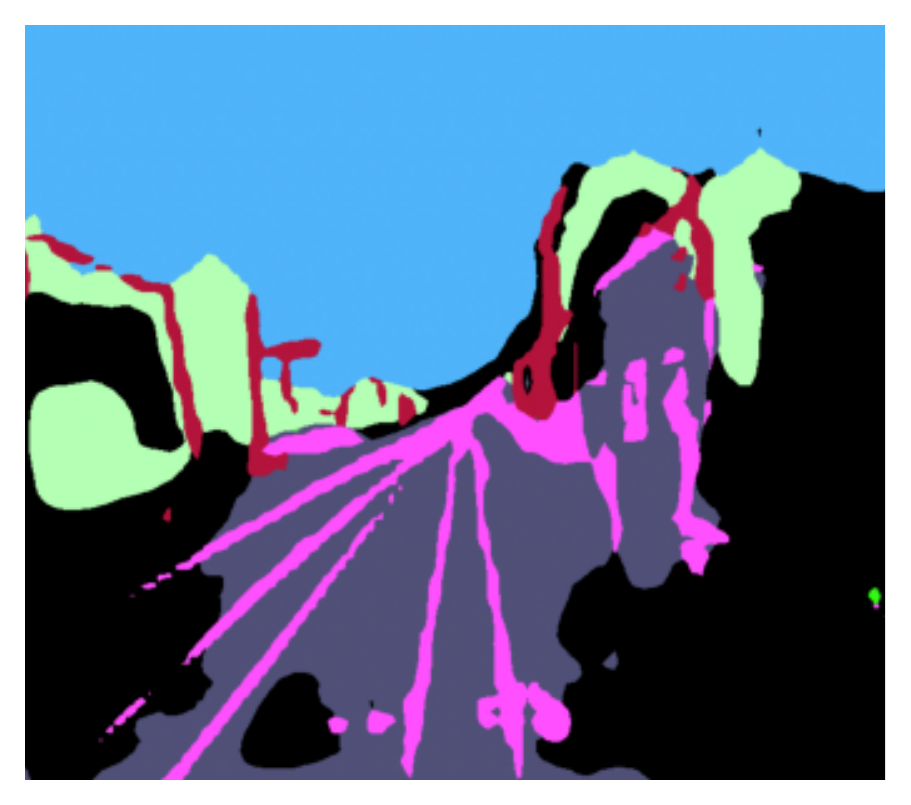}
     \end{subfigure}
     \begin{subfigure}{0.32\columnwidth}
         \centering
         \includegraphics[width=\textwidth]{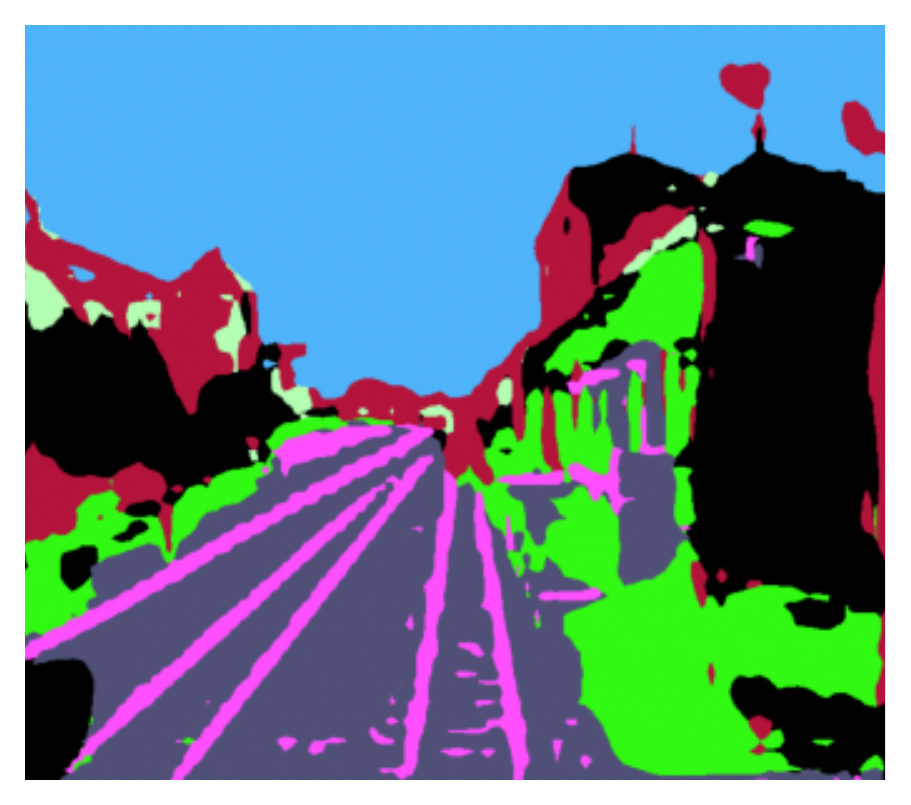}
     \end{subfigure}
     \begin{subfigure}{0.32\columnwidth}
         \centering
         \includegraphics[width=\textwidth]{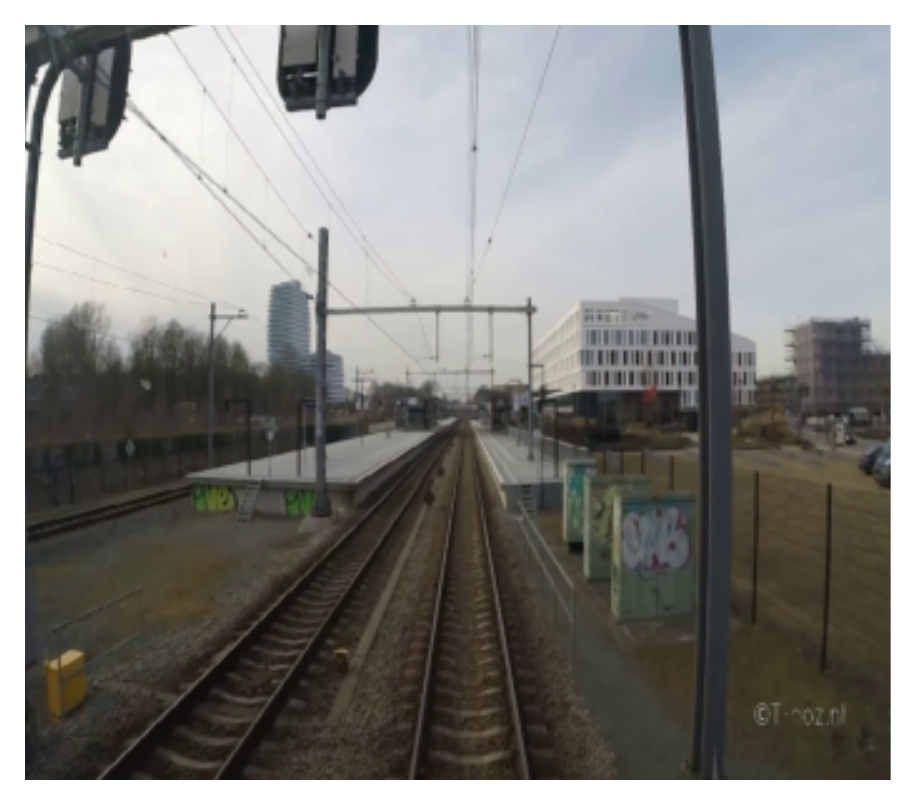}
         \caption{input}
     \end{subfigure}
     \begin{subfigure}{0.32\columnwidth}
         \centering
         \includegraphics[width=\textwidth]{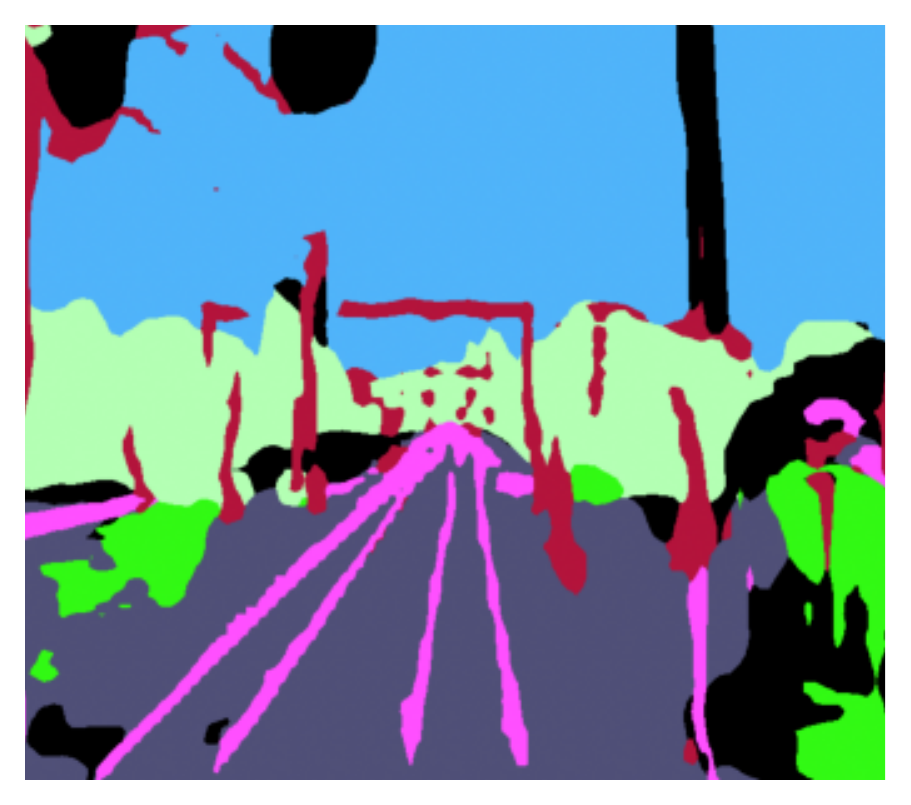}
          \caption{SS-20}
     \end{subfigure}
     \begin{subfigure}{0.32\columnwidth}
         \centering
         \includegraphics[width=\textwidth]{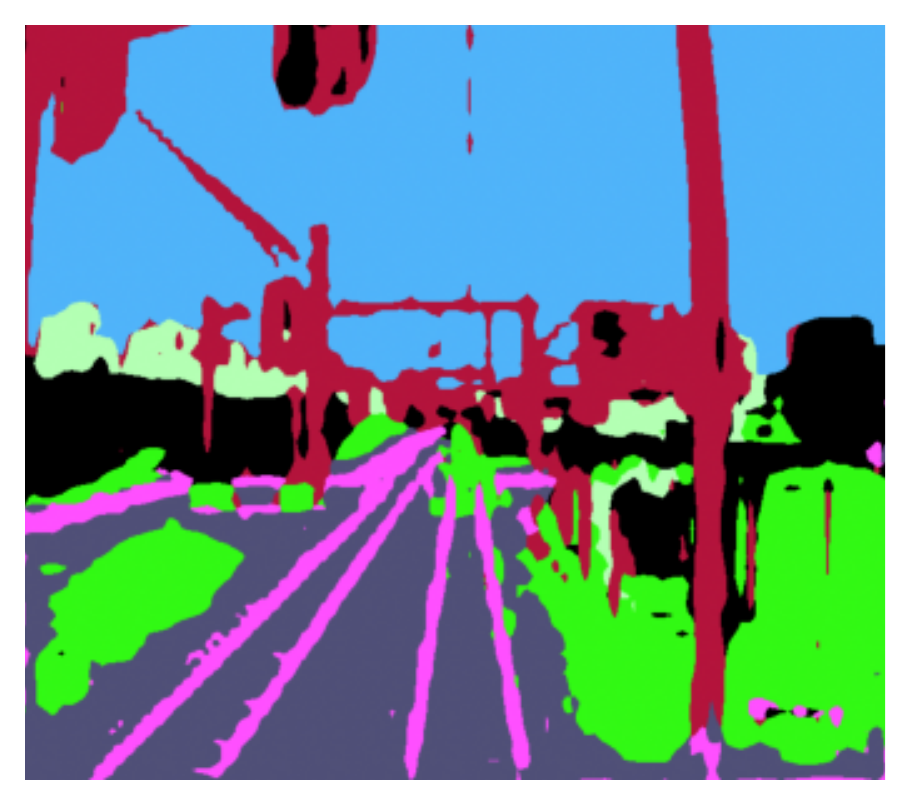}
         \caption{SSDA-20}
     \end{subfigure}
    \caption{\small{Output predictions of RailSem real-world images.}}
    \label{fig:output_bisenet}
\end{figure}

In SSDA mode, the training set was augmented with 6700 annotated synthetic images collected from different simulated scenarios, similar to those described in Section~\ref{ss:lidarOdomA}, where different materials were randomly applied to the trackbed and the landscapes, and various lighting conditions were used to add some variability to the gathered images. Since RailSem and TrainSim define two different sets of object classes, the analysis was conducted on a subset of RailSem classes also present in TrainSim (see Table~\ref{table:table_SS} and Figure~\ref{fig:output_bisenet}), while all the remaining classes were considered as `background'.

The neural architecture selected for the semantic segmentation task is a BiseNetX39 \cite{bisenet}, trained by the Adam optimizer \cite{adam} with its default settings and a learning rate of 0.003.
Batch size and training steps were set to 30 and 8000, respectively. The training was performed by using the classic pixel-wise cross-entropy loss. Input images were resized to ($H$=680, $W$=720) to reduce the computational cost, while random crop (scale $1/2$) and random horizontal flip were used for training set augmentation.

Table~\ref{table:table_SS} reports the performance achieved on the RailSem dataset (details in the caption), showing that the use of TrainSim improves the IoU on crucial classes (rail-track, trackbed, and terrain), whereas the performance on other classes is reduced, most likely due to a more accentuated domain shift between synthetic and real-world textures.
Figure~\ref{fig:output_bisenet} shows two real-world images taken from RailSem (a) and the corresponding segmented images produced by SS-20 (b) and SSDA-20 (c).
In accordance with Table~\ref{table:table_SS}, the model trained using synthetic samples (SSDA-20) produces more accurate segmentation maps.

\begin{table}
\centering
\setlength\tabcolsep{5pt}
\begin{tabular}{l|cc|cc}
Class &  SS-10 &  SSDA-10 &  SS-20 &  SSDA-20 \\
\hline
pole       &         \textbf{0.104} {\footnotesize $\pm$0.11} &      0.095 {\footnotesize $\pm$0.13} &                        \textbf{0.117} {\footnotesize $\pm$0.16}&               0.102 {\footnotesize $\pm$0.11} \\
vegetation &         \textbf{0.344} {\footnotesize $\pm$0.26} &      0.298 {\footnotesize $\pm$0.79} &                        \textbf{0.407} {\footnotesize $\pm$0.17} &               0.366 {\footnotesize $\pm$0.26}\\
terrain    &         0.102 {\footnotesize $\pm$0.78}&       \textbf{0.156} {\footnotesize $\pm$0.14} &                       0.164 {\footnotesize $\pm$0.35}&               \textbf{0.194} {\footnotesize $\pm$0.16} \\
sky        &         0.740 {\footnotesize $\pm$0.67}&       0.740 {\footnotesize $\pm$0.10} &                       0.773 {\footnotesize $\pm$0.13}&               \textbf{0.780} {\footnotesize $\pm$0.14} \\
trackbed   &         0.325 {\footnotesize $\pm$0.14} &      \textbf{0.366} {\footnotesize $\pm$0.41} &                        0.391 {\footnotesize $\pm$0.44}&               \textbf{0.415} {\footnotesize $\pm$0.05}\\
rail-track &         0.175 {\footnotesize $\pm$0.06}&       \textbf{0.192} {\footnotesize $\pm$0.18} &                       0.197 {\footnotesize $\pm$0.26}&               \textbf{0.208} {\footnotesize $\pm$0.22} 
\end{tabular}
\caption{\small{Performance of the BiseNet model \cite{bisenet} achieved by a semi-supervised (SS) approach (with only RailSem samples) and a semi-supervised domain adaptation (SSDA) with both real-world and synthetic samples (RailSem + TrainSim). The values denote the Intersection over Union (IoU) and $std\times10$ of each class among a 4-fold cross-validation on RailSem. `10' and `20' denote the number of real-world annotated samples, randomly extracted from RailSem.}}
\label{table:table_SS}
\end{table}

Despite the benefits discussed above, we also noticed that increasing the number of annotated real-world samples (i.e., more than 50) SSDA yields lower performance than SS (without TrainSim images).
We believe this is due to a semantic domain shift between the simulated and real-world images, which become more relevant for a higher number of real-world samples. For instance, TrainSim does not account for complex textures contained in RailSem (e.g., crowded urban and driving scenarios). This forces the model to learn a constrained subset of visual patterns during SSDA, forgetting those that are not well-represented in TrainSim but still useful in real-world scenarios.
Please also note that such a domain shift is more accentuated when running an unsupervised DA or without any DA strategy. This motivated us to explore SSDA, where a small subset of real-world data helps alleviate the domain shift.

These points open interesting future works for investigating novel DA approaches for railway scenarios.
Finally, it is also worth remarking that, to the best of our knowledge, this work is the first one that proposes an SSDA approach for semantic segmentation in railway scenarios.

\section{Conclusions} \label{s:conclusions}

This paper presented TrainSim, a visual simulation framework designed to automatically generate a number of realistic railway scenarios and produce labeled datasets from emulated sensors, as LiDARs, cameras, and inertial measurement units. 
Such datasets are exported in a format suitable for training deep neural models for object detection, semantic segmentation, and depth estimation for camera data, or for processing 3D point clouds from a LiDAR.
For each 3D point, the LiDAR model also provides the intensity of the backscattered ray, which can be used to simplify the discrimination of the tracks from other objects with higher diffusion coefficients.

The preliminary experiments carried out on the simulated sensors show the effectiveness of the proposed approach, making the simulator a useful tool for investigating and testing new perception algorithms for railway applications.

As a future work, we plan to extend the simulator by adding railway switches and meshes with higher fidelity, and upgrading TrainSim to Unreal Engine 5 to exploit its newer features, such as improved photorealism.
Finally, the results reported in Section~\ref{ss:imageSemA} on image segmentation provide interesting insights for further investigating the use of domain adaptation in railway environments.

\bibliographystyle{IEEEtran}
\bibliography{rel.bib}

\vfill

\end{document}